\documentclass[11pt]{article} % use larger type; default would be 10pt
\usepackage{geometry} % to change the page dimensions
\usepackage{booktabs} % for much better looking tables
\usepackage{array} % for better arrays (eg matrices) in maths
\usepackage{paralist} % very flexible & customisable lists (eg. enumerate/itemize, etc.)
\usepackage{verbatim} % adds environment for commenting out blocks of text & for better verbatim
\usepackage{amssymb}
\usepackage{amsmath}
\usepackage{graphicx}
\usepackage{epstopdf}
\usepackage{subcaption}
\usepackage[round, sort, comma, authoryear]{natbib}
\usepackage{float}
\usepackage{authblk}
\usepackage{tikz}
\usetikzlibrary{arrows,shapes}
\usepackage{algorithm}
\usepackage{algpseudocode}
\usepackage{tikz}
\usetikzlibrary{arrows,shapes}
\usepackage{arydshln}

\title{A tutorial on recursive models for analyzing and predicting path choice behavior}
\author[1,2]{Ma\"elle Zimmermann, Emma Frejinger}

\affil[1]{Department of Computer Science and Operations Research, Universit\'e de Montr\'eal, QC, Canada}
\affil[2]{CIRRELT - Interuniversity Research Centre on Entreprise Networks, Logistics and Transportation, Montr\'eal, QC, Canada}

\begin{document}

\maketitle
\begin{abstract}
    The problem at the heart of this tutorial consists in modeling the path choice behavior of network users. This problem has been extensively studied in transportation science, where it is known as the route choice problem. In this literature, individuals' choice of paths are typically predicted using discrete choice models. 
    This article is a tutorial on a specific category of discrete choice models called recursive, and it makes three main contributions: First, for the purpose of assisting future research on route choice, we provide a comprehensive background on the problem, linking it to different fields including inverse optimization and inverse reinforcement learning. Second, we formally introduce the problem and the recursive modeling idea along with an overview of existing models, their properties and applications. Third, we extensively analyze illustrative examples from different angles so that a novice reader can gain intuition on the problem and the advantages provided by recursive models in comparison to path-based ones. 
    %The aim of this tutorial is to present this problem from the novel and more general perspective of inverse optimization, in order to describe the modeling approaches proposed in related research areas and thereby motivate the use of so-called recursive models. The latter have the advantage of predicting path choices without generating choice sets. In this paper, we contextualize discrete choice models as a probabilistic approach to an inverse shortest path problem with noisy data, highlighting that recursive discrete choice models in particular originate from viewing the inner shortest path problem as a parametric Markov Decision Process. We also illustrate through simple numerical examples that recursive models overcome issues associated with the path-based discrete choice models commonly found in the transportation literature.

\end{abstract}

\section{Introduction}

Road traffic, while essential to the proper functioning of a city, generates a number of nuisances, including pollution, noise, delays and accidents. It is the role of city managers, network administrators and urban planners to attempt to mitigate the negative impact of transportation by planning adequate infrastructure and policies. Most of the traffic is generated by individual travelers who seek to minimize their own travel costs without guidance from a system maximizing the overall welfare. It is thus a necessity to understand how users of the transportation system behave and choose their path in the network in order to provide planners with decision-aid tools to manage it.

This is the main motivation for the introduction of what is known in the transport demand modeling literature as the route choice problem, which seeks to predict and explain travelers' path choices in a network. The path choice models we attend to are based on the assumption that individuals behave rationally by minimizing a certain cost function or, equivalently, maximizing a certain utility function. This function can depend on network attributes and possibly socio-economic characteristics of the travellers'. The models' aim is to identify this cost function from a set of observed trajectories (revealed preferences data), which allows to predict chosen paths for all origin-destination (OD) pairs. All models in this tutorial are based on the assumption that network attributes are deterministic (static or dynamic). We refer the reader to, e.g., \cite{GaoFrejBenA10} for the case of stochastic attributes.

Path choice predictions are rarely useful on their own. Rather, path choice models are used in combination with information about aggregate demand given by an OD matrix. The latter states the number of travellers who wish to travel between each OD pair in a given time interval. The problem of computing traffic flows given an OD matrix, constant travel times and a path choice model is called stochastic traffic assignment. In fact, the key idea behind recursive path choice models -- avoiding explicit path enumeration -- originates from the traffic assignment model in \cite{Dial71}. In turn, the resulting traffic flows are a central component in a number of problems with broad applicability. A prominent example is the computation of equilibrium flows when travel times are given as a function of flow \citep[see, e.g.,][for recursive formulations]{BELL1995287,HuanBell98,akamatsu1996cyclic,baillon2008markovian}. Traffic flows are also crucial to network design problems \citep{MagnWong84} and bilevel optimization formulations thereof \citep[e.g.,][]{BrotEtAl08,GilbertEtAl15}.
The literature on each of these problems is rich, and a detailed description is out of the scope of this tutorial that is entirely devoted to disaggregate recursive path choice models.

%Path predictions are in general not the ultimate goal, and instead form a central component of equilibrium, assignment or network design models.

With the aforementioned problems in mind, it is desirable that path choice models satisfy several properties, such as (i) scaling with the size of large urban networks in order to be efficiently used for real networks, (ii) lending themselves well to behavioral interpretation, in order to, e.g., assess travelers' value of time, and (iii) yielding accurate predictions under different network scenarios.
In the transportation demand literature, the most common methodology is a probabilistic approach known as discrete choice modeling, which finds its origin in econometrics. It views path choice as a particular demand modeling problem, where the alternatives of the decision maker are the paths in the network. The principal issue with this modeling approach stems from the fact that the set of feasible paths in a real network cannot be enumerated, and we do not observe which paths are actually considered by individuals. As a result, discrete choice models based on paths currently used in the transportation literature may suffer from biased estimates and inaccurate predictions, as well as potentially long computational times.

This work is a tutorial on a modeling framework which in comparison meets the previously enumerated expectations. Recursive route choice models draw their efficiency from modeling the path choice problem as a parametric Markov decision process (MDP) and resorting to dynamic programming to solve its embedded shortest path problem.
% This paragraph has to answer the questions: to whom is the tutorial addressed? What is in its scope and outside the scope? Explain that it is not a classical tutorial and spends somewhat more time giving a comprehensive background and making important connections to other fields. Then explain the actual aims.
This tutorial is addressed to both transportation demand researchers who are unfamiliar with recursive models as well as researchers and practitioners who wish to gain understanding of the route choice problem and how it can be modelled. For this reason, we do not assume from the reader knowledge in discrete choice modeling and instead provide a new and comprehensive background on the route choice problem (which for generality purposes we denote path choice from here on). Rather than presenting recursive models in their entirety, we aim at deepening their conceptual understanding, in particular i) their link to inverse shortest path and inverse reinforcement learning problems, ii) their advantages over the most well-known approach in the transportation demand literature, i.e., discrete choice models based on paths, iii) guidance related to their applicability to large-scale problems. %know-how %limitations

% PREVIOUS TEXT: In this tutorial, we aim at i) deepening understanding of recursive models' formulation by drawing links to related work in inverse optimization, and ii) comparing recursive models on the basis of desirable properties with the most well-known approach in the transportation demand literature, i.e., discrete choice models based on paths. We note here that this tutorial is addressed to both transportation demand researchers who are unfamiliar with recursive models, and researchers from the machine learning community who are keen to find out about state-of-the-art methods in the area of transportation science. For this purpose, we use general terminology and speak about path choice instead of route choice in the remaining of this paper.

% Perhaps better to elaborate on the three parts idea in the abstract and detail the importance and contribution of each
This tutorial is essentially divided into three parts. The first gives an extensive background on the problem it addresses and makes important connections to other fields in Section \ref{secContext}. We contextualize discrete choice models as a probabilistic approach for what is in fact an inverse shortest path problem with noisy data. The second part, in Section \ref{secprobamodels}, formally introduces path-based and recursive route choice models and provides a review of the different recursive model types with references to technical papers. It also provides a review of existing applications supplied by guidance on scaling the models to larger networks. The third part consists of illustrative examples for educational purposes: Section \ref{secexample} provides a comparison of two simple recursive and path-based discrete choice models on toy networks for the illustration of basic concepts and properties for the uninitiated reader. Section \ref{secchoicesets} discusses in depth the issues related to estimation and prediction and demonstrates the advantages of recursive models through practical examples and discussions. Finally, Section \ref{secconclu} provides an outlook and concludes.

\section{Context: from shortest paths problems to path choice models}
\label{secContext}
In this section, we frame the path choice problem as that of unveiling an unknown cost function from noisy shortest paths observations, which is an inverse optimization problem where the forward (inner) problem is a shortest path. We give some background on the literature on inverse optimization and we situate the problem this tutorial addresses within this context. We illustrate that there is a close connection between stochastic shortest path problems and the inverse problem with noisy data we are interested in. For the sake of clarity we start this section by introducing deterministic and stochastic shortest path problems before describing the related inverse problems.

\subsection{Shortest path problems}
\label{secshortestpath}
Throughout this section, we consider a simple oriented graph $G$ with a set of nodes $\mathcal{V}$ and a set of arcs $\mathcal{A}=\{(i,j) \ | \ i, j \in \mathcal{V}\}$. We denote $v$ the nodes in $\mathcal{V}$ and $a$ the arcs in $\mathcal{A}$, which are characterized by a source node $i_a$ and a tail node $j_a$. Arcs $(i,j)$ have an associated cost $c_{ij}$ given by a function $c: \mathcal{A} \rightarrow \mathbb{R}$. A path is a sequence of arcs such that the head node of each arc is the tail node of the next. 

\subsubsection{The deterministic shortest path problem}
\label{secdetshortest}
The \emph{deterministic shortest path} problem (DSP) in the graph $G$ is concerned with finding the path with minimum cost between an origin node and a destination node, where the cost of a path is defined by the sum of its arc costs. More often, methods developed in the literature are designed to solve the shortest path problem between a given origin and all possible destinations, or a given destination and all possible origins. 

This combinatorial optimization problem has been amply studied in the literature. Its chief difficulty is the existence of a very large number of feasible paths between each node pair, which precludes proceeding by naive enumeration. The problem could be formulated and solved as a linear program, however more efficient algorithms have been developed, relying on \emph{dynamic programming} (DP). In general, DP is a methodology to solve optimization problems in dynamic (often discrete time) systems, where a decision (denoted action or control) must be taken in each state in order to minimize future additive costs over a certain time horizon (finite or infinite). The shortest path problem in the graph $G$ can be formulated as a DP problem by considering nodes as states and an arc choice as an action taken in a given state.

The Bellman principle of optimality at the core of deterministic problems states that for an optimal sequence of choices (in this case, arcs along the shortest path), each subsequence is also be optimal. This allows to decompose the problem and formulate a recursive expression for the optimal arc choice at node $i$ as well as the cost $C(i)$ of the shortest path from $i$ to destination node $d$, 

\begin{equation}
    \label{Bellman}
        C(i) = \left\{
		\begin{array}{ll}
		 0, & i = d\\
         \min_{j\in\mathcal{V}_i^+}\{c_{ij}+C(j)\},& \forall i \neq d, i\in \mathcal{V}
        \end{array} \right.
\end{equation}
where $\mathcal{V}_i^+ = \{j \in \mathcal{V}\ |\ (i,j)\in \mathcal{A}\}$.%where $c_{ij}$ denotes the cost of the arc $(i,j)$. %This is known as the Bellman equation, and the quantity $C$ is also known as the value function in dynamic programming. 

Solving (\ref{Bellman}) is however not straightforward in cyclic graphs. In this case, note that the problem is well-defined only when there are no negative cost cycles, otherwise there would be paths of cost $-\infty$. Under this assumption, the shortest path in $G$ contains at most $|\mathcal{V}|-1$ arcs. \cite{bellman1958routing} shows how to solve the shortest path problem by backwards induction as a deterministic finite state finite horizon optimal control problem. The DP algorithm is 
\begin{equation}
    \label{Bellman2}
        \left\{
		\begin{array}{ll}
		 C_N(i) = c_{id}, & \forall i \in \mathcal{V}_d^-\\
         C_n(i) = \min_{j\in \mathcal{V}_i^+}\{c_{ij}+C_{n+1}(j)\},& \forall i \in \mathcal{V}, n=1,...,N-1
        \end{array} \right.
\end{equation}
where $\mathcal{V}_i^- = \{h \in \mathcal{V}\ |\ (h,i)\in \mathcal{A}\}$, and $N = |\mathcal{V}|-1$ is the length of the horizon. The value $C_n(i)$ represents the cost of the shortest path from $i$ to $d$ using at most $N-n+1$ arcs, and in this sense is an upper bound on the cost of the shortest path. The cost of the shortest path from $i$ corresponds to $C_1(i)$. Note that the value of the shortest paths can also be found by label correction algorithms \citep[see, e.g.,][]{dijkstra1959note,floyd1962algorithm}. 

\subsubsection{The stochastic shortest path problem}
\label{secstochshortest}
The \emph{stochastic shortest path} problem (SSP), as defined by \cite{bertsekas1991analysis}, is an extension of the previous problem which considers a discrete time dynamic system where a decision must be taken in each state and causes the system to move stochastically to a new state according to a transition probability distribution. This problem can be analyzed using the framework of Markovian Decision Processes (MDP), formally defined as:
\begin{itemize}
    \item A set of states $S$ and a set of available actions $A(s)$ for each state $s\in S$.
    \item The cost $c_{s,a,s'}$ incurred by taking action $a$ in state $s$ and moving to next state $s'$.
    \item{$p(s'|s,a)$ the transition probability from $s$ to $s'$ when taking action $a$.}
\end{itemize}
An MDP models problems where an action must be taken in each state, with the aim to minimize \emph{expected} future discounted costs over a certain horizon. The SSP is a special case of MDP with infinite horizon, no discounting and a cost-free absorbing state $d$, where $p(d|d,a)=1 \ \forall a$. The SSP is an infinite horizon problem since there is no upper limit on the number of arcs traversed. However, by assumption the absorbing state can be reached with probability 1 in finite time. The optimal solution of the SSP is not a path but a policy, which consists in a probability distribution over all possible actions in each state. When the optimal policy is followed, the path which is actually travelled is random but has minimum expected cost. 

%The SSP is modeled as a particular MDP in which the destination is an absorbing state $d$ from which no costs are endured and $p(d|d,a)=1$ for all available actions $a$. This results in the absorbing state being reached with probability 1 in finite time. In contrast to classic infinite horizon MDPs, costs are not discounted over time. 
This stochastic problem can be solved with DP and the recursion which defines the optimal expected cost $V(s)$ from any given state $s$ is given by the Bellman equation
\begin{equation}
\label{BellmanSSP}
    V(s) = \min_{a\in A(s)} \left[ c_{s,a} +\sum_{s'} p(s'|s,a) V(s')\right],
\end{equation}
where $c_{s,a} = \sum_{s'} p(s'|s,a)c_{s,a,s'}$. $V(s)$ is also known as the \emph{value function}. 

Note that the DSP is a particular degenerate case of the SSP where states $s$ are nodes $i$ of the graph $G$, action costs are arc costs $c_{ij}$ and state transition probabilities are deterministic, since the next state is always equal to the chosen successor node. Therefore, it is in fact a deterministic MDP.

This definition of SSPs provided by \cite{bertsekas1991analysis} is very general. It also encompasses variants of shortest paths problems on random graphs \citep[see, e.g.,][]{polychronopoulos1996stochastic}, where arc costs are modeled as random variables $\tilde{c}_{ij} = c_{ij} +  \varepsilon_{ij}$, with $c_{ij}$ the average arc cost and $\varepsilon_{ij}$ a random error term. Arguably the most interesting variation of the problem is the one where the realization of the arc costs is learned at each intersection as the graph is traversed. Below, we make the additional assumption that $\varepsilon_{ij}$ are i.i.d. variables.
In this case, by defining an action $a$ as an arc $(i,j)$ and a state $s$ as a network node $i$ and a vector $e_i$ of learned realizations $e_{ij}$ of the error term for all outgoing arcs from $i$, (\ref{BellmanSSP}) becomes 
%\textcolor{red}{Take the point of view of the traveller. Who observes the realization of epsilon at each intersection before taking action. This is additional information at each stage.}

%V(i,e_i) = \min_{j\in \mathcal{V}_i^+} \left[ c_{ij}+ e_{ij} + \int_{\epsilon_j} V(j,e_j) f(\epsilon_j) \right],

\begin{equation}
\label{BellmanSSP2}
    V(i,e_i) = \min_{j\in \mathcal{V}_i^+} \left[ c_{ij}+ e_{ij} + \int V(j,e_j) f(de_j) \right],
\end{equation}
where transitions between states are entirely contained by the density $f(e_j)$. This specific SSP will be of importance in Section \ref{TheRL}.

\subsection{Inverse shortest path problems}
In shortest path problems it is assumed that the modeler has complete knowledge of the cost function $c$ or its distribution in the previous case, as well as state transitions $p$. In contrast, \emph{inverse shortest path problems} study the case where the cost function is unknown and must be inferred with the help of an additional source of information at disposal, in the form of observed optimal paths between some origin-destination (OD) pairs. This class of problems broadly belongs to inverse optimization, an extensively studied problem \citep[see e.g.][]{ahuja2001inverse} where the modeler seeks to infer the objective function (and sometimes the constraints) of a forward (inner) optimization problem based on a sample of optimal solutions. 

Motivation for studying inverse problems can be drawn from several types of applications. \cite{burton1992instance} cites possible motivations to examine inverse shortest path problems, one of which is precisely the subject of this tutorial. One may view the underlying optimization problem as a model for rational human decision making and assume that the cost function represents the preferences of users traveling in a network (possibly a parametrized function of certain arc features). In this context, recovering the cost function allows to analyze why individuals choose the observed routes and to gain understanding of network users' behavior. Using the recovered cost function, the inner shortest path problem can be solved and in this sense yield predictions of path choices for unobserved OD pairs. 

%Note that we are interested in the \emph{deterministic} inverse shortest path problem, which assumes that network users do not face cost uncertainties or stochastic state transitions linked to unpredictable events. On the contrary, the traveler is assumed to have complete knowledge of the network. The stochastic counterpart (the inverse of Section \ref{secstochshortest}'s problem) has been little studied, to the exception of \cite{zhou2014inverse}.

Related to inverse shortest path problems is the literature on \emph{inverse reinforcement learning} (IRL) or imitation learning \citep{ng2000algorithms, abbeel2004apprenticeship}. The IRL problem is more general than the inverse shortest path, since it considers an underlying optimization problem generally formulated as a infinite horizon MDP. In this context, observations consist of optimal sequences of actions. Nevertheless, models for IRL have also been applied to the problem of recovering the cost function of network users \citep[e.g.,][]{ziebart2008maximum}. Such applications consider specific MDPs where an action is a choice of arc in a network, and the destination is an absorbing state, as in Section \ref{secstochshortest}. Most applications of IRL on path choice however consider a deterministic MDP (state transitions probabilities are degenerate) in contrast to the more general formulation in Section \ref{secstochshortest}.

Inverse problems are in general under-determined and may not possess a unique solution. In the inverse shortest path problem and in IRL, there may be an infinite number of ways to define the cost function such that observations form optimal solutions. Different modeling paradigms propose to solve this issue. They may be separated in two categories depending on the assumptions made on the presence of noise in the data, which distinguishes between deterministic and stochastic problems. The former consider that demonstrated behavior is optimal, while the latter makes the hypothesis that observed trajectories deviate from deterministic shortest paths. The second case is often studied when the data collecting process may have induced measurement errors or when the data is generated by a decision maker who exhibits seemingly inconsistent behavior. In the following sections, we review existing models for the above mentioned inverse problems.

\subsubsection{Deterministic problems}

\cite{burton1992instance} introduced the original deterministic inverse shortest path problem where observed trajectories are assumed to correspond to deterministic shortest paths. They do not assume that the cost function is parametrized by arc features and simply seek the value $c_{ij}$ associated to each arc $(i,j)$. They propose to provide uniqueness to the inverse problem by seeking the arc costs $c$ that are closest to a given estimation $\hat{c}$ of costs based on a certain measure of distance, thus minimizing an objective of the form $||c-\hat{c}||$. This implies that the modeler has an a priori knowledge of costs, which is reasonable in some applications. \cite{burton1992instance} provide seismic tomography as an example. Seismic waves are known to propagate according to the shortest path along the Earth's crust, but the geological structure of the zone of study is typically not entirely known, although modelers have an estimate. Given measurements of earthquakes' arrival times at different points in the ``network'', the goal is then to predict movements of future earthquakes by recovering the actual transmission times of seismic waves.

Variants of this problem have been studied by \cite{burton1994use,burton1997inverse}, for instance considering the case where arcs costs may be correlated or their values belong to a certain range. In \cite{burton1992instance} and following works, the $l_2$ norm is selected as a distance measure between initial and modified costs, and yields a quadratic programming formulation. In \cite{zhang1995column}, the $l_1$ norm is assumed so that the problem can be modelled as a linear program and solved using column generation. In all the studies above, the inverse shortest past problem is modeled as a constrained optimization problem, with an exponential number of linear constraints to ensure that each observed path is shortest under the chosen cost function. 

\cite{barmann2017emulating} provide another example of approach for inverse optimization with an application to learning the travel costs of network users, in particular subject to budget constraints. More precisely, they consider an inverse resource-constrained shortest path problem. Their approach does not recover an exact cost function, but provides a sequence of cost functions $(c_1,...,c_T)$ corresponding to each observation $t=1,...,T$, which allows to replicate demonstrated behavior. Their framework explicitly assumes the optimality of observations in order to infer the objective functions $c_t$.

The deterministic problem has also notably been studied under the guise of inverse reinforcement learning in \cite{ratliff2006maximum}, with an application to robot path planning. The objective of their work is to learn a cost function in order to teach a robot to imitate observed trajectories, which are assumed optimal. In contrast to the aforementioned works, no hypothesis is made regarding preliminary arc costs. However, the environment is considered to be described by features, such as elevation, slope or presence of vegetation, and the model seeks to obtain a mapping from features to costs by learning the weights associated to each feature. To obtain uniqueness of the solution, they cast the problem as one of maximum margin planning, i.e., choosing the parameters of the cost function that makes observed trajectories better by a certain margin than any other path, while minimizing the norm of weights. This notion of distance between solutions is defined by a loss function to be determined by the modeler. Under the $l_2$ norm, this also results in a quadratic programming formulation with a number of constraints that depend on the number of state-action pairs, which consist of node-arc pairs in this context.

\subsubsection{Stochastic problems}
Inverse optimization problems with noisy data have been studied in, e.g., \cite{aswani2018inverse} or \cite{chan2018inverse}. In addition to non-uniqueness, the problem which typically arises in this situation is that there may not be any non trivial value of arc costs which makes the demonstrated paths optimal solutions of a DSP. If solutions do exist, they may be uninformative, such as the zero cost function. To solve this issue, the previous framework is extended by letting solutions be approximately optimal and measuring the amount of error. Thus accommodating noise requires to estimate a model for the choice of path which ``fits'' as closely as possible the observed data with various methods for measuring the fit, or loss, drawing from statistics. The chosen measure for the fit should uniquely define the solution.

Different points of views exist on achieving a good fit in the literature. Approaches grounded in machine learning make no assumptions on the underlying process that generated the data and merely focus on obtaining good predictive power with the simplest possible model while considering a large family of potential functions. In contrast, the statistical inference perspective on the problem considers that there exists an underlying true cost function with a known parametric form. The aim is to obtain parameter estimates that asymptotically converge to the real values, a property known as \emph{consistency}. Several loss functions are conceivable to formulate a minimization problem, and often the choice of loss is directly related to assumptions made on the underlying model. 
%and that observed behavior deviates randomly from optimality according to a certain probability distribution. without making assumptions regarding distribution of the observations.

Examples for the former are the work of \cite{keshavarz2011imputing}, who estimate cost functions in a flow network assuming an affine parametric shape, and the work of \cite{bertsimas2015data}, who seek to infer cost functions in a network subject to congestion at equilibrium. The specificity of the latter is that the cost functions are endogenous, i.e., the cost of paths include congestion costs. This makes it an inverse variational inequality problem with noisy data. In both works the proposed method is a heuristic and treats the process that generates the data as a black box. For example, in \cite{bertsimas2015data}, nonparametric cost functions are considered and the problem is formulated as a constraint programming model which balances the objective of minimizing the norm of the cost function with that of maximizing the fit of the data. They follow the approach of measuring the loss of the model by the amount of slack required to accommodate equilibrium constraints, i.e., making observed solutions ``$\epsilon$-optimal''. 

In contrast, the latter category of models assume that observed behavior deviates randomly from optimality according to a certain known probability distribution. This leads to a different modeling paradigm, in which a random term is added to the true cost function of the inner optimization problem, with the interpretation that each observation corresponds to an instantiation of the cost. This framework, described in, e.g., \cite{nielsen2004learning} for general inverse problems, has received limited attention in the literature on inverse shortest path problems. One notable exception consists in path choice models based on the discrete choice modeling framework. In the next section, we review the literature surrounding this probabilistic modeling framework, which is at the heart of this tutorial.

\section{Probabilistic models for path choice}
\label{secprobamodels}
%The inverse shortest path problem with noisy data is motivated by the situation of individuals traveling in real networks, where the assumption that a cost function can account for all observed behavior is rarely valid. 
Individuals traveling in real networks is a typical situation where the assumption that a cost function can account for all observed behavior is rarely valid. Probabilistic models for this inverse problem with noisy data assume that observed path choices randomly deviate from deterministic shortest paths. In particular, discrete choice models are a particular type of probabilistic models grounded in econometrics, which provides the theoretical basis for behavioral interpretation. 

One of the interpretations of the distributional assumption on the data is that travelers act rationally but observe additional factors impacting their path choice which vary among individuals and are unknown to the modeler. These factors are encompassed in a random term $\varepsilon$ added to the cost function. Discrete choice models are based on the assumption that the cost function is a parametrized function of attributes. The only option available to the modeler, who knows the family of distributions for $\varepsilon$ but does not observe the realization of random terms for a given individual, is to infer the probability that a given path is optimal from data. 

The problem becomes akin to density estimation, i.e., recovering the parameters of a probability distribution over a set of paths. In this context, statistical consistency of the estimator is a desirable property. Yet there are several ways to define a probability distribution over paths, well-known in the discrete choice literature, which do not necessarily yield a consistent estimator. In the following sections, we elaborate on the above statement and describe two distinct categories of discrete choice models for path choice. 

Note that discrete choice models employ the terminology of utilities instead of costs and that we uphold this convention in the remainder of this tutorial. This implies a trivial change of the above formulations from minimization to maximization problems and the definition of a utility function $v=-c$.

\subsection{Path-based models}
\label{secprobmodels}
Discrete choice models based on paths are the methodology embraced by most works on the topic \citep[e.g.,][]{prato2009route, frejinger2008route}. They assume that the utility of a path $i$ is a random variable $u_i = v_i + \mu\varepsilon_i$, where $\varepsilon_i$ is a random error term, $\mu$ is its scale, and the deterministic utility $v_i$ is parametrized by attributes of the network, such as travel time. Often the parametrization consists of a linear-in-parameters formulation $v_i = \beta^Tx_i$, where $\beta$ is a vector of parameters and $x_i$ is a vector of attribute variables of the path $i$. The parameter values are inferred from observed path choices $\{\sigma_n\}_{n=1,...,N}$ through maximum likelihood estimation, as we further detail below.
%Measuring the fit of a discrete choice model to the data naturally leads to selecting the log-likelihood loss, which assesses the plausibility of observing the chosen trajectories $\{\sigma_n\}_{n=1,...,N}$ under the current value of parameters $\beta$. Thus the problem is one of maximum likelihood estimation, i.e., finding the set of parameters which minimize the log-likelihood loss.

A key difficulty in specifying the probability of a given path is to identify the set of paths over which this probability should be defined. Since the very large number of feasible paths in a real network precludes enumeration, the immediate solution consists in choosing a subset of reasonable paths, assuming that all other paths have a null choice probability. This implies a two-step modeling framework \citep{Manski1977}: the probability of choosing path $\sigma_n$ is $P(\sigma_n)=\sum_{C_n \in G_n} P(\sigma_n|C_n)P(C_n|G_n)$ where $C_n$ is a choice set of paths and $G_n$ is the set of all non-empty subsets of the set of all feasible paths. This expression hence involves an exponential sum that is impossible to evaluate in a path choice context as there is already an exponential number of feasible paths. Therefore, most path-based models ignore the sum and consider one unique subset of path $C_n$ for each $n$, i.e., assuming that $P(\sigma_n)=P(\sigma_n|C_n)$. The models are defined and estimated in the following manner:
\begin{enumerate}
    \item Plausible paths are generated between the origin and destination of each observation $n$ by solving versions of the DSP, forming the choice set $C_n$ (a vast array of methods exist and we refer to \cite{bekhor2006evaluation} for an overview);
    \item Parameters $\hat{\beta}$ that maximize the probability $P(\sigma_n|C_n; \beta)$ of observed paths within the choice set previously defined are estimated via maximum likelihood, i.e., by maximizing $\mathcal{LL}(\beta)$ defined as 
    \begin{equation}
        \label{likelihoodPL}
        \mathcal{LL}(\beta) = \log \sum_{n=1}^N P(\sigma_n|C_n;\beta).
    \end{equation}
\end{enumerate}
The distribution chosen for $\varepsilon_i$ leads to different types of discrete choice models. The most well-known is the multinomial logit formulation, resulting from the assumption of i.i.d. Extreme value type I distributed error terms, also known as the softmax in the machine learning community 
\begin{equation}
    \label{logit}
    P(\sigma_n|C_n;\beta)=\frac{e^{\frac{1}{\mu}v_{\sigma_n}(\beta)}}{\sum_{j\in C_n}e^{\frac{1}{\mu}v_{jn}(\beta)}}.
\end{equation}
Many other models exist, such as the link-nested logit, mixed logit and probit models \citep{vovsha1998link, yai1997multinomial, frejinger2007}. 

%There exists a vast array of methods to extract plausible paths from the network in order to generate the needed choice sets. Usually, it consists in assuming an a priori cost function and solving variants of the DSP described in Section \ref{secshortestpath}, until a large enough set of paths is obtained. Problematically, the preliminary value of costs used to define the probability distribution through the choice set is in general not equivalent to the true value of the utility which is ultimately sought. This discrepancy is what prevents the consistency of the resulting estimates. 
The first step in the two-step approach may be viewed as deterministic or stochastic sampling of alternatives, depending on the algorithm. Problematically, the costs used for the DSP do in general not correspond to the utilities which are ultimately sought. This discrepancy is what prevents the consistency of the resulting estimates. However, if the utilities are adequately corrected to account for to the sampling protocol, then the estimator is still consistent \citep{mcfadden1978}. 
\cite{frejinger2009} propose this idea for path choice modeling and delineate a method that corrects the 
%to correct for the induced sampling bias, by adjusting the 
%choice probability $P(\sigma_n|C_n; \beta)$ of a given path depending on parameter values and the choice set. The adjusted 
path choice probabilities for the probability $P(C_n | \sigma_n)$ that the given choice set was selected conditionally on the observed choice. 

The method proposed by \cite{frejinger2009} can be understood as assuming that the true distribution is based on the set consisting of all feasible paths, while resorting to sampling paths in order to estimate the parameters of the distribution in practice. While the sampling correction allows to consistently estimate path-based models, it is still unclear how to use them for prediction. Designing a sampling correction in this context is an open problem. We further discuss this issue in Section~\ref{secpredict}.
%The advantage is that it yields consistent parameter estimates. 
%Nevertheless, since there is no means to compute the normalizing constant of the distribution save for the impractical enumeration of all paths, the estimated model still requires to sample choice sets for prediction, an issue we further discuss in Section \ref{secpredict}.

% Ce n'est pas le IRL qui provide une solution, en fait la key idee est d'utiliser le DP pour modeliser le inner shortest path problems. Un example ou cela est fait est en IRL. Arguably, just like DP provides a solution for shortest path problems, it can also provide one for inverse problems.
Arguably, since these issues arise from the exponential number of feasible paths, one may expect that DP could provide a solution for the inverse problem as well.
The literature on IRL supplies such an example with \cite{ziebart2008maximum}, who model the path choice problem as a MDP and estimate a probabilistic model which normalizes over the global set of feasible paths. This is achieved without enumeration nor sampling by viewing the path choice process as a sequence of action choices depending on a current state as in Section \ref{secstochshortest}. %Although they employ the principle of maximum entropy to justify their framework, it is equivalent to a maximum likelihood estimation problem.  
In fact, this methodology is similar to recursive discrete choice models, which have been proposed independently and in parallel in the transportation research community. We introduce this modeling framework in the following section, which provides a method to consistently estimate parameters of the utility function without resorting to the two-step modeling approach with choice sets of paths.

\subsection{Recursive models}
\label{TheRL}
In this section, we introduce recursive models for the problem of path choice within the discrete choice modeling framework. We first present the initial model proposed by \cite{fosgerau2013}, before reviewing its subsequent extensions. We also discuss the link to the IRL model by \cite{ziebart2008maximum}.

\subsubsection{Path choice as a deterministic MDP}
Contrary to the previous section in which the problem is formulated in the high dimensional space of paths, recursive models consider arc-based variables. Their formulation is based on the framework of MDP used to solve shortest path problems in Section \ref{secshortestpath}.

In recursive models, network arcs correspond to states, while outgoing links at the head node of the current arc assume the role of available actions. For this purpose, we subsequently denote arcs as either $k$ or $a$ depending on whether they play the role of states or actions. Note that it would also be possible to select nodes as states, however, the arc-based formulation allows the deterministic utility $v(a|k)$ of an action-state pair to depend on turn angles between two subsequent arcs $k$ and $a$. The destination is represented by a dummy link $d$ which is an absorbing state of the MDP, where no additional utility is gained. Finally, utilities are undiscounted and the action-state transition function is assumed to be degenerate, since the new state is simply the chosen arc. A path under this framework is a sequence of states $\{k_0,k_1,...,k_T\}$, starting from an origin state $k_0$ and leading to the absorbing state $k_T$ representing the destination $d$.

\subsubsection{Estimation of the parametric MDP}
Under this perspective, the inverse problem of recovering the utility function is a problem of parametric estimation of MDPs, as first described by \cite{rust1994structural}.
As in the previous section, the noise in the data is accounted for by assuming the presence of a random error term $\varepsilon_a$, added to the utility $v(a|k)$. The utility becomes a random variable $u(a|k) = v(a|k) + \mu\varepsilon_a$, where $\mu$ is the scale of the error term. It is assumed that the individual observes the realizations of the random variables at each step of the process and chooses the best action accordingly. From the point of view of the modeler, the individual's behavior hence consists in solving a stochastic shortest path problem similarly to Section \ref{secstochshortest}. In this context, the Bellman equation gives the optimal value function when the state consists of an arc $k$ and realizations $e_{a\in \mathcal{A}(k)}$ of $\varepsilon_{a\in \mathcal{A}(k)}$,
\begin{equation} \label{Bellmanepsilon}
    V^d(k,e_{a})= \left \{ 
	\begin{array}{ll}
	 0, & k=d,\\
    \max_{a\in A(k)}\left(v(a|k)+\mu e_a+\int V^d(a,e_{a'}) f(de_{a'}) \right), & \forall k\in \mathcal{A},
    \end{array} \right. 
\end{equation}
which is very similar to (\ref{BellmanSSP2}). We may however simplify this equation by taking the expectation with respect to $\varepsilon_a$ of (\ref{Bellmanepsilon}) and defining the expected value function $V^d(k) = \int V^d(k,e_a) f(de_a)$ of a state $k$, which gives

\begin{equation}
    \label{Bellmanepsilon2}
    V^d(k)=\left\{
	\begin{array}{ll}
	 0, & k=d,\\
     E_{\epsilon}\left[\max_{a\in A(k)}\left\{v(a|k)+\mu\epsilon_a+V^d(a)\right\}\right], & \forall k\in \mathcal{A}.
    \end{array} \right.
\end{equation}
For simplicity and consistency with terminology in other works \citep[e.g.,][]{fosgerau2013,mai2015nested}, we nevertheless refer to (\ref{Bellmanepsilon2}) as the value function in this work.

The modeler does not observe the realized utilities and can only compute the probability that each given action is optimal. According to the modeler, the observed behavior of individuals follows a probability distribution over the set of actions which maximizes the expected utility in (\ref{Bellmanepsilon2}). As in Section \ref{secprobmodels}, choice probabilities may take several forms depending on the distributional assumption for the error terms $\epsilon_a$. When the modeler assumes that they are i.i.d. Extreme value type I, the model is known as the recursive logit (RL) model \citep{fosgerau2013}. In Section \ref{existingmodels}, we provide an overview of existing models that relax the i.i.d. assumption.

Under the RL model, the probability of an individual choosing a certain action $a$ conditional on the state $k$ and the destination $d$ is given by the familiar multinomial logit formula:

\begin{equation}
\label{equlinkprob}
	P^d(a|k;\beta)=\frac{e^{\frac{1}{\mu}v(a|k;\beta)+V^d(a|\beta)}}{\sum_{a'\in A(k)} e^{\frac{1}{\mu}v(a'|k;\beta)+V^d(a'|\beta)}}.
\end{equation}

Given observations of sequences of actions (i.e., paths), the model can be estimated by maximum likelihood. This requires to define the probability of choosing an observed path, which can be expressed as the product of the corresponding action choice probabilities using (\ref{equlinkprob}). 
For an observed path $\sigma_n=\{k_0,k_1,...,k_T\}$, the path choice probability is given by
\begin{equation}
P(\sigma_n|\beta)=\prod_{j=0}^{T-1}P^d(k_{j+1}|k_j ;\beta ),
\end{equation}
where $d$ is the link $k_T$, which can more simply be expressed as
\begin{equation}
P(\sigma_n|\beta)=\frac{e^{\frac{1}{\mu}v(\sigma_n|\beta)}}{e^{\frac{1}{\mu}V(k_0| \beta )}}
\end{equation}
where $v(\sigma_n|\beta)$ is the sum of the link utilities of the path $\sigma_n$.

As a result, the likelihood of a set of $N$ path observations $\{\sigma_n\}_{n=1,...,N}$ is defined as
\begin{equation}
\label{likelihood}
	\mathcal{L}(\beta)=\prod_{n=1}^N P(\sigma_n|\beta) = \prod_{n=1}^N \prod_{i=0}^{T_n-1} P^d(k^n_{i+1}|k^n_{i};\beta).
\end{equation}

The expression in (\ref{likelihood}) does not depend on choice sets, in contrast to (\ref{likelihoodPL}). However, the value function which appears in (\ref{equlinkprob}) must be computed in order to evaluate the likelihood. This suggests resorting to a two-step likelihood maximization algorithm, in which an inner loop solves the SSP and obtains the value function in (\ref{Bellmanepsilon2}) for the current value of parameters $\beta$, while an outer loop searches over the parameter space. \cite{rust1994structural} propose such a method, denoted the \emph{nested fixed point} (NFXP) algorithm. The resulting parameter estimates are consistent. We note that alternative estimators to the nested fixed point algorithm exist, but have not yet been used in the context of path choice modeling \citep[see, e.g.,][]{aguirregabiria2010dynamic, SuJudd12}.

The model for IRL proposed in \cite{ziebart2008maximum} bears another name but is equivalent to the RL model, since they assume a maximum entropy (exponential family) distribution. The difference lies in the method for estimating the model, as \cite{ziebart2008maximum} approximate the value function in (\ref{Bellmanepsilon2}), whereas \cite{fosgerau2013, mai2015decomposition} show that they can be conveniently solved as a system of linear equations without approximation. We provide an illustration of the latter on small examples in Section~\ref{secexample}.
%that they can conveniently be solved as a system of linear equations \citep{fosgerau2013, mai2015decomposition}.

\subsubsection{Existing recursive models and their applications}
\label{existingmodels}
Path choice modelling entails two key challenges. First, the definition of the choice sets, or more precisely, the support for the choice probability distribution. Second, given the network structure, paths likely share unobserved attributes which lead to correlated random terms. 
%While inverse reinforcement learning models fail to consider the latter, 
It has been shown that accounting for this correlation is crucial to achieve accurate predictions and interpretation of parameter estimates. The RL model addresses the first challenge but not the second as the random terms are assumed to be i.i.d. This section is devoted to recursive models that have been designed to relax this assumption and hence address also the second challenge. We provide an overview of their properties and related applications. For a more in-depth description of each model, we refer to the corresponding articles.

In the context of recursive models it is important to note that changing the distributional assumption of the random terms $\varepsilon_a$, changes the expression of the value functions (\ref{Bellmanepsilon2}). When one relaxes the i.i.d. Extreme Value Type I assumption, the value functions are solutions to non-linear systems instead of linear ones, which are harder to solve. As the computation of the value functions constitutes the inner loop of the nested fixed point algorithm, even a slight increase in the computing time might lead to an important increase in the overall computing time associated with estimating the model parameters. The estimation also depends on the outer non-linear optimization algorithm. There are many possible algorithms that can be used for this purpose. As the estimation typically is done offline (i.e., the models are not re-estimated during prediction) high computing times may be acceptable, of course, within reasonable limits. Nevertheless, computing the value functions as fast as possible to a good accuracy is also crucial for prediction. 

We report in Table~\ref{table_models} a list of existing models. For each model we provide a reference to the article where it is introduced, we state whether the choice probabilities have a closed form (CF) or not, and if the model allows utilities to be correlated (Corr) or not. Moreover, we report in the second last column how the value functions were computed and the non-linear optimization algorithm that was used in each of the articles. A key point is the ability to model correlated utilities. In this regard, more modeling flexibility results in higher computational time as a trade-off. The cross-nested (RCNL) model is more flexible than the nested (NRL) but the computation of the value functions is more complex. The mixed RL (MRL) can, in theory, approximate any correlation structure but requires a computationally intensive simulation approach. 

Some methods are available to reduce the computational burden associated with more complex models. For example, the time required to estimate the MRL is reduced by a decomposition method \citep[DeC, introduced in ][]{mai2015decomposition} and the use of common random numbers. The DeC method allows to solve one system of linear equations for all destinations in the case of RL, which results in an important speed up. In order to solve value functions which are solutions to non-linear systems, it is necessary to resort to value iteration methods that are widely applicable but can be slow to converge. This is why several works use a warm start (an initialisation using a RL solution computed by solving systems of linear equations) and dynamic accuracy. The latter means that the solutions are only computed with a high accuracy close to the maximum likelihood estimates. 

\begin{table}
    \centering
    \fontsize{8pt}{10.2}\selectfont
    \begin{tabular}{p{4cm} p{3cm} c c p{3cm} c}
    Model & Reference & CF$^{\text{(1)}}$ & Corr$^{\text{(2)}}$ & Value functions & Opt$^{\text{(3)}}$ \\
    \toprule
    Recursive logit (RL) & \cite{fosgerau2013} & Y  & N & Linear system of equations (decomposition) & BFGS\\
    Maximum entropy IRL & \cite{ziebart2008maximum} & Y & N & VI$^{\text{(4)}}$ inspired algorithm & NR$^{\text{(6)}}$\\ 
    Nested recursive logit (NRL) & \cite{mai2015nested} & Y & Y & Warm-started VI$^{\text{(4)}}$ and dynamic accuracy & BFGS \\
    Mixed recursive logit (MRL) & \cite{mai2015decomposition} & N & Y & Decomposition, common random numbers & BFGS \\
    Recursive cross-nested logit (RCNL) & \cite{mai2016method} & Y & Y & Warm-started VI$^{\text{(4)}}$ and dynamic accuracy & BFGS \\
    %Recursive network MEV (RNMEV) & \cite{mai2016method} & \\
    Random regret-based recursive logit (RRM-RL) & \cite{mai2017similarities} & Y & Y & Linear system of equations & IP$^{\text{(5)}}$ \\
    Discounted recursive logit (DRL) & \cite{oyama2017discounted} & Y & N & VI$^{\text{(4)}}$  & NR$^{\text{(6)}}$ \\
    %Maybe add strategy-based recursive logit? by Nassir? & \\
    \toprule
    \multicolumn{2}{l}{(1) Closed-form probability} & \\
    \multicolumn{2}{l}{(2) Correlated utilities} & \\
    \multicolumn{2}{l}{(3) Precision on non-linear optimization algorithm}  & \\
    \multicolumn{2}{l}{(4) Value iteration (method of successive approximations)} & \\
    \multicolumn{2}{l}{(5) Interior point method} \\
    \multicolumn{2}{l}{(6) Not reported in the article.} \\
    \end{tabular}
    \caption{Information on existing recursive route choice models}
    \label{table_models}
\end{table}

Next we turn our attention to the applications of these models on real networks, in order to show how the computation time increases for larger network sizes and to present the available methods to deal with large-scale challenges. In Table~\ref{table_applis} we summarize a number of articles reporting the type of application, the number of links in the graph (i.e., number of states of the MDP), the model, the number of estimated parameters as well as the reported computing times. We note that one has to be careful when analyzing these computing times as they depend on the data (in particular size of the dataset), choice of algorithms, their implementation as well as hardware. We report them to give an indication on the orders of magnitude. Arguably, it would have been more informative to present the computing time at the level of value functions (averages and the distribution over the destinations for each network). These values are however not reported across the different studies and there are even studies not mentioning computing times at all. The reported figures show clearly that the computing times are considerably longer for models that account for correlated utilities \citep[compare for example, for the same data, the RL and NRL in the bike route choice application by][]{zimmermann2017bike}. On the other hand when the utilities are discounted \citep{oyama2017discounted}, the utility interpretation changes but the model has nice properties as the value iteration converges faster and to a unique solution (contraction mapping).

Computing times depend not only on the type of model being applied, but also on the type of graph and estimation method. Acyclic graphs, such as time-expanded networks, are advantageous for the computation of value functions, as they can be solved quickly by backwards induction (a process we illustrate in Section \ref{secexample}), resulting in gains of time for both estimation and prediction. In terms of estimation method, sampling of alternatives represents a possible substitute when the nested fixed point algorithm is too costly. Although this method was initially designed for path-based models in \cite{frejinger2009}, it can also be applied to some recursive models and possesses several additional advantages in this case. Compared to the NFXP, sampling of alternative obtains large gains in estimation time, but importantly, it remains possible to predict path choices using the estimated link choice probabilities without generating choice sets. \cite{zimmermann2018transit} provide an example of the application of sampling of alternatives to estimate the parameters of a MRL model on the largest graph in terms of size in Table~\ref{table_applis}. Note that sampling of alternatives yields unbiased estimates when used in conjunction with RL and MRL models, however these results have not yet been extended to more complex recursive models.
%\cite{zimmermann2018transit} report the largest application in terms of size of the graph. In addition, this is a dense graph of activities as opposed to a sparse graph of a road network. For graphs of this size, it would be too costly to use the nested fixed point estimator. Instead, this work uses a sampling of alternatives approach to estimate the parameters of a MRL model which allows to use advantages of the recursive model for prediction.

Another aspect to consider in order to use recursive models in practice are requirements in terms of utility specification. Recursive models need link-additive utilities, which can be challenging to specify when inherently non-additive attributes, such as the slope of a road, are likely to have an impact on path choice. We note that \cite{zimmermann2017bike} deals with such challenges in the context of a bike route choice application. In some applications, it may also be desirable to include socio-economic characteristics of individuals in the utility specification, as in \cite{zimmermann2018capturing}. It is important to note that in this case, value functions need to be solved not only for each destination, but also for each category of individuals defined in the utilities. 

\begin{table}
    \centering
    \fontsize{8pt}{10.2}\selectfont
    \begin{tabular}{p{3.5cm} p{2.5cm} c c c c r}
    Reference & Application & Acyclic & Nb links & Model & Nb param. & Est. time [h]\\
    \toprule
    \cite{ziebart2008maximum} & Taxi route choice & N & 300,000 & IRL & NR$^{\text{(1)}}$ & NR$^{\text{(1)}}$ \\
    \cite{zimmermann2017bike} & Bike route choice & N &42,384 & RL & 14& 1 \\
    & & & & NRL &15 & 336  \\
    \cite{de2019modelling} & Multimodal route choice & N &127,873 & RL & 12 & 96\\
    \cite{zimmermann2018capturing}& Activity choice & Y & millions & MRL & 48 & 3 \\
    \cite{nassir2018strategy}& Transit route choice & N & NR$^{\text{(1)}}$ & RL & 7 & NR$^{\text{(1)}}$ \\
    \cite{oyama2017discounted}& Car route choice & N & 2724 & DRL & 3 & 0.08 \\
    \cite{mai2015nested}& Car route choice & N & 7459 & NRL & 7 & 30 \\
    \cite{mai2016method}& Car route choice & N & 7459 & RCNL & 8 & 72 \\
    \cite{mai2015decomposition} & Car route choice & N & 7459 & MRL & 5-10 & 72-120 \\
    \toprule
    \multicolumn{2}{l}{(1) Not reported in the article.} & \\
    \end{tabular}
    \caption{Overview of applications of recursive route choice models}
    \label{table_applis}
\end{table}

\section{Illustrative examples}  
\label{secexample}
Section \ref{secprobamodels} presented two families of discrete choice models for the problem of estimating the utility function of travelers in a network: path-based models and recursive models.
Although the former is extensively used in practice, the latter is superior because of its consistent estimator and accurate predictions without choice set generation.

This section is addressed to readers who are new to the topic and provides small illustrative examples on toy networks with two purposes in mind. First, we provide an intuitive understanding of the value function in recursive models. Second, we compare path probabilities obtained with both model formulations in order to highlight a key property of recursive models, namely, that the recursive logit model is equivalent to a path-based logit over the complete set of possible paths. Although both models can take several forms depending on distributional assumptions on the error terms, the focus of these illustrative examples is on the logit model. In the following, we refer to the recursive logit as the RL model and the path-based logit as the PL model.

Note that the MATLAB code used for these numerical examples is available online, as well as a tutorial detailing how to use it\footnote{ http://intermodal.iro.umontreal.ca/software.html}. 

\subsection{An acyclic network}
\begin{figure}
	\begin{center}
		\begin{tikzpicture}[scale = 0.8]
		\tikzstyle{sommet}=[circle,draw,thick,fill=white]
		\node[sommet] (1)at(0,0){$1$};
		\node[sommet] (2)at(3,0){$2$};
		\node[sommet] (4)at(0,-4){$4$};
		\node[sommet] (5)at(3,-3){$3$};
		\draw [dashed,->,>=latex] (0,1.5) -- (1);
		\draw [->,>=latex](1) -- node[midway, left] {$2$}(4);
		\draw [->,>=latex](1) -- node[midway, above] {$1$}(2);
		\draw [->,>=latex](2) -- node[midway, below right] {$2$}(4);
		\draw [->,>=latex](2) -- node[midway, right] {$1.5$} (5);
		\draw [->,>=latex](5) to[bend left] node[midway, below right] {$1.5$}(4);
        \draw [->,>=latex](1) to[out=200,in=160] node[midway, left] {$6$}(4);
        \draw [dashed,->,>=latex] (4) -- (0,-5.5);
		\end{tikzpicture}
	\end{center}
	\caption{Small network}
	\label{fignetwork}
\end{figure}
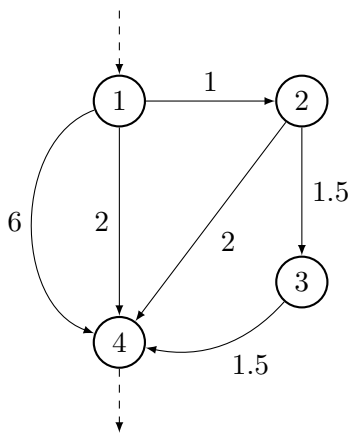

The motivation for this example is to show that it is possible to obtain with the recursive formulation in (\ref{equlinkprob}) the same choice probabilities on paths as with the PL model in (\ref{logit}). For this illustrative purpose, it is meaningful to consider a given specification of the utility function. Hence we assume that path utilities are specified by an additive function of arc length $L_a$, such that for each path $i$ we have $v_i=-L_i$, where $L_i$ is the sum of the lengths of arcs contained in path $i$.

The OD pair considered for the toy network displayed in Figure \ref{fignetwork} is $(1,4)$. The two dashed arcs represent dummy origin and destination links. There exists 4 possible paths from node 1 to node 4, of respective length 2, 3, 4 and 6. Under the logit model, it is easy to compute the choice probability of the shortest path for this OD pair, which goes directly from node 1 to node 4 with length 2. Assuming that the scale $\mu$ of the random term for this example is 1, we obtain at the denominator of the logit function in (\ref{logit}) the term $e^{-2}$, and at the numerator the term $e^{-2}+e^{-3}+e^{-4}+e^{-6}$, therefore the choice probability is equal to 0.6572. Table \ref{tablogit} displays similarly the choice probability of all other paths.

Let us now suppose that instead of choosing between the possible paths connecting origin and destination, the traveler builds the chosen path along the way through a series of consecutive link choices, as in the RL formulation. In each link choice situation, the alternatives to choose from are the outgoing links at the current node. We denote $v(a|k) = -L_a$ the utility of links $a\in\mathcal{A}(k)$ originating from link $k$. The choice probability of a path under the RL model is then equal to the product of each link choice probability in (\ref{equlinkprob}).

\begin{table}
	 	\begin{center}
	 		\begin{tabular}{lccr}
	 			Path & Length & Path probability (PL) & Product of link probabilities (RL) \\
	 			\midrule
	 			$1-4$ & 2& 0.6572 & 0.6572\\
	 			$1-4$ & 6& 0.0120 & 0.0120\\
	 			$1-2-4$ & 3& 0.2418 & $0.3307\cdot 0.7311=0.2418$\\
                $1-2-3-4$ & 4& 0.0889 & $0.3307\cdot 0.2689=0.0889$
	 		\end{tabular}
	 		\caption{Path choice probabilities under both models}
	 		\label{tablogit}
	 	\end{center}
	 \end{table}
     \begin{table}
	 	\begin{center}
	 		\begin{tabular}{cc}
	 			Node & Value function \\
	 			\midrule
	 			$4$ & $0$  \\
	 			$3$ & $-1.5000$ \\
	 			$2$ & $-1.6867$\\
                $1$ & $-1.5803$ 
	 		\end{tabular}
	 		\caption{Value function at each node}
	 		\label{tabvaluefunction}
	 	\end{center}
	 \end{table}
	 
 In order to compute link choice probabilities, we need to compute the value function in (\ref{Bellmanepsilon}). This equation can be rewritten as the logsum when $\epsilon_a$ is assumed to be i.i.d. Extreme value type I distributed as in the logit model:
\begin{equation}
\label{logsum}
V^d(k)=\left\{
\begin{array}{ll}
\mu\ln{\sum_{a\in A(k)}e^{\frac{1}{\mu}(v(a|k)+V^d(a))}} & \forall k \in \mathcal{A}\\
0 & k=d.
\end{array} \right.
\end{equation}
Since the specified utility of a link does not depend on the incoming arc, the value function is identical for all links with the same end node. It is thus more convenient to compute the value function at each of the 4 nodes (the value function at a link is then equivalent to the value function at the end node of that link). Below, we show how to compute the value function in this network and display the value for each node.

In this case, since the network is acyclic, it is possible to compute the value function by backwards induction. At the destination node 4, given that there is no utility to be gained, the value function $V(4)$ is zero. Working our way backwards, we compute at node 3 the value function $V(3) = \ln(e^{-1.5}) = 1.5$. At node 2, we have $V(2)=\ln(e^{-2}+e^{-3}) = -1.6867$. Finally, at node 1 we obtain $V(1) = \ln(e^{-6}+ e^{-2}+ e^{-2.6867}) = -1.5803$.
The values for all nodes are summarized in Table \ref{tabvaluefunction}.

Having computed the value function for this network, we may apply (\ref{equlinkprob}) to this example and we obtain the path probabilities in the last column of Table \ref{tablogit}. We notice that they are identical to choice probabilities under the PL model. This is due to the property of the RL model of being formally equivalent to a discrete choice model over the full choice set of paths \citep{fosgerau2013}. Therefore, the PL and the RL models are two strictly equivalent approaches when the set of all possible paths in the network can be enumerated.

 \subsection{A cyclic network}
\begin{figure}
	\begin{center}
		\begin{tikzpicture}[scale = 0.8]
		\tikzstyle{sommet}=[circle,draw,thick,fill=white]
		\node[sommet] (1)at(0,0){$1$};
		\node[sommet] (2)at(3,0){$2$};
		\node[sommet] (4)at(0,-4){$4$};
		\node[sommet] (5)at(3,-3){$3$};
		\draw [dashed,->,>=latex] (0,1.5) -- (1);
		\draw [->,>=latex](1) -- node[midway, left] {$2$}(4);
		\draw [->,>=latex](1) -- node[midway, above] {$1$}(2);
        \draw [->,>=latex](5) -- node[midway, above] {$1$}(1);
		\draw [->,>=latex](2) -- node[midway, below right] {$2$}(4);
		\draw [->,>=latex](2) -- node[midway, right] {$1.5$} (5);
		\draw [->,>=latex](5) to[bend left] node[midway, below right] {$1.5$}(4);
        \draw [->,>=latex](1) to[out=200,in=160] node[midway, left] {$6$}(4);
        \draw [dashed,->,>=latex] (4) -- (0,-5.5);
		\end{tikzpicture}
	\end{center}
	\caption{Small cyclic network}
	\label{fignetwork2}
\end{figure}
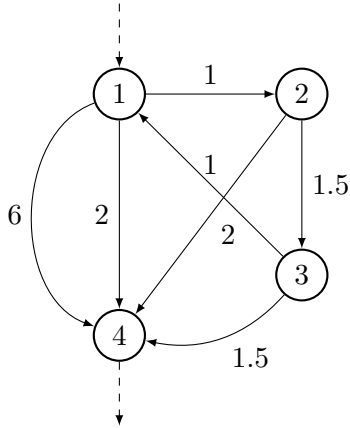

 Let us now consider a very similar network in Figure \ref{fignetwork2}, with an added link between nodes 3 and 1. This network is no longer acyclic, and as a result there is in theory an infinite number of paths between nodes 1 and 4, when accounting for paths with loops.
 
 The first consequence of dealing with a cyclic network is that the value function can no longer be computed by backwards induction starting from destination, since the network admits no topological order. However, the value function is still well defined as the solution of the fixed point problem (\ref{Bellmanepsilon}) and can be solved either by value iteration or, in the case of the recursive logit, as the solution of a system of linear equations. For the latter, notice that by taking the exponential of (\ref{logsum}) and raising to the power $\frac{1}{\mu}$, we obtain
\begin{equation}
\label{linearizedV}
e^{\frac{1}{\mu}V^d(k)}=\left\{
\begin{array}{ll}
\sum_{a\in A(k)}\delta(a|k)e^{\frac{1}{\mu}(v(a|k)+V^d(a))} & \forall k \in \mathcal{A}\\
1 & k=d.
\end{array} \right.
\end{equation}
 This is a linear system of equations if we solve for variable $z=e^{\frac{1}{\mu}V}$. Doing so, we obtain the value function in Table \ref{tabvaluefunction2}.
 
 As in the previous example, having solved the value function, we can trivially compute the choice probabilities for different paths in this network as product of link choice probabilities. 
 As can be observed, the probabilities of the four paths used in the acyclic example do not sum to 1 anymore (rather to 0.9698), and neither do the probabilities of the additional paths displayed in Table \ref{tabreclogit2}, which sum to 0.9965. This is because a cyclic network contains an infinite number of possible paths, and the RL model attributes a positive choice probability to each outgoing arc at an intersection. Hence, even paths with multiple cycles have a small probability of being chosen. We notice however that choosing a path with two or more cycles is extremely unlikely, with a probability of 0.0009 according to the model.
 
 \begin{table}
 	\begin{center}
 		\begin{tabular}{cc}
 			Node & Value function \\
 			\midrule
 			$4$ & $0$  \\
 			$3$ & $-1.1998$ \\
 			$2$ & $-1.5968$\\
            $1$ & $-1.5496$ 
 		\end{tabular}
 		\caption{Value function at each node}
 		\label{tabvaluefunction2}
 	\end{center}
 \end{table}
 \begin{table}
 	\begin{center}
 		\begin{tabular}{lcr}
 			Path & Length & Product of link choice probabilities \\
 			\midrule
 			$1-4$ & 2& 0.6374 \\
 			$1-4$ & 6& 0.0117 \\
 			$1-2-4$ & 3& $0.3509\cdot 0.6682=0.2345$ \\
            $1-2-3-4$ & 4 & $0.3509\cdot 0.3318 \cdot 0.7407=0.0863$\\
            $1-2-3-1-4$& 5.5 & $0.3509\cdot 0.3318 \cdot 0.2593\cdot 0.6374=0.0192$\\
            $1-2-3-1-4$& 9.5 & $0.3509\cdot 0.3318 \cdot 0.2593\cdot 0.0117=0.0004$\\
            $1-2-3-1-2-4$& 6.5 & $0.3509\cdot 0.3318 \cdot 0.2593\cdot 0.3509\cdot 0.6682= 0.0071$
 		\end{tabular}
 		\caption{Recursive logit path choice probabilities}
 		\label{tabreclogit2}
 	\end{center}
 \end{table}
 
This example illustrates that the RL model offers a convenient mathematical formulation for the choice of path in a cyclic network. In comparison, using the PL model for this network raise a well-known challenge. Indeed, the logit formula in (\ref{logit}) requires to define a finite choice set of alternative paths for the OD pair. Given that the possible paths cannot be all enumerated in this example, the modeler is compelled to make hypotheses on which subset of paths should have a non zero choice probability. The value of the resulting path choice probabilities will depend the composition of the choice set. In reality, this issue is not necessarily related to cycles only. In large networks, the number of possible acyclic paths may also be too large to enumerate in practice. In the following section, we delve into the issues which may arise from the necessity to generate choice sets to define choice probabilities in path-based models.
 
\section{An analysis of the advantages of recursive models compared to path-based models}
\label{secchoicesets}
The goal of this section is to highlight in more detail the advantages of recursive models and the issues related to path-based models. In this discussion, we use illustrative examples and we focus on two practical purposes of such models; i) estimating parameters from data of observed paths; ii) predicting choices from an estimated model. We focus on logit models for this comparison.

In practice, the PL model requires to generate choice sets of paths for both purposes. There is an extensive literature on the questions of how to generate choice sets of paths, what characteristics should choice sets observe, and what is the impact of selecting a restricted choice set prior to model estimation and prediction \citep{bekhor2006evaluation,prato2007modeling,bovy2009modelling,bliemer2008impact}. The consensus in that literature is that it is advantageous to explicitly separate the procedures of generating path choice sets and modelling choice. \cite{bekhor2005investigating} argue that predicted paths from link-based models are behaviorally unrealistic as they may contain cycles. \cite{bliemer2006route} claim that there are computational advantages to choice set generation in large networks. 

On the contrary, \cite{horowitz1995role} indicate that it is possible to mis specify choice sets with problematic consequences and that choice sets provide no information on preferences besides what is already contained in the utility function, although their study does not investigate path choice. \cite{frejinger2009}, among others, empirically demonstrate that the definition of choice sets may affect parameter estimates. In this section of this tutorial, we offer additional arguments in this sense. We exemplify complications related to choice sets which arise when using path-based models, and we demonstrate that recursive logit models do not display these issues.

%There is however little research on whether consideration sets should be altogether avoided in route choice modeling, and the impact of using the universal set of feasible paths. Given that this is now a possibility with the RL model and in light of the practical difficulties of identifying true consideration sets in practice, the questions which the remainder of this paper seeks to answer are
% \begin{enumerate}
% 	\item Can a model based on an ill-chosen restricted choice set have biased estimates and make erroneous predictions?
% 	\item Can including irrelevant alternatives in the choice set result in the model having biased estimates and making erroneous predictions?
%    \item Are there practical advantages in using the universal set of feasible paths for either estimation or prediction?
% \end{enumerate}
 
\subsection{An example of model estimation}
\label{secexample1}

 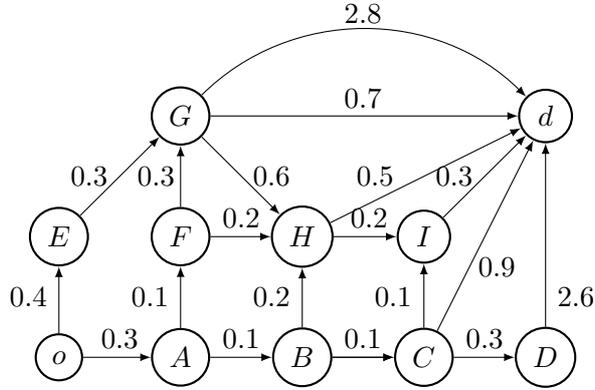
\begin{figure}
	\begin{center}
		\begin{tikzpicture}[scale=0.8]
		%\node (legend) at (2.5,-3){Recursive logit};
		\tikzstyle{sommet}=[circle,draw,thick,fill=white]
		\node[sommet] (1)at(0,0){$o$};
		\node[sommet] (2)at(2,0){$A$};
		\node[sommet] (3)at(4,0){$B$};
		\node[sommet] (4)at(6,0){$C$};
		\node[sommet] (5)at(8,0){$D$};
		\node[sommet] (6)at(0,2){$E$};
		\node[sommet] (7)at(2,2){$F$};
		\node[sommet] (8)at(2,4){$G$};
		\node[sommet] (9)at(4,2){$H$};
		\node[sommet] (10)at(6,2){$I$};
		\node[sommet] (11)at(8,4){$d$};
		%\draw (0,0) node[scale=1]{$\bullet$} node[left]{$O$} ;
		%\draw (4,0) node[scale=1]{$\bullet$} node[right]{$D$} ;
		
		\draw[->,>=latex](1)--(2);
		\draw[->,>=latex] (2) -- (3);
		\draw[->,>=latex] (3) -- (4);
		\draw[->,>=latex] (4) -- (5);
		\draw[->,>=latex] (1) -- (6);
		\draw[->,>=latex] (2) -- (7);
		\draw[->,>=latex] (3) -- (4);
		\draw[->,>=latex] (7) -- (8);
		\draw[->,>=latex] (7) -- (9);
		\draw[->,>=latex] (9) -- (10);
		\draw[->,>=latex] (3) -- (9);
		\draw[->,>=latex] (4) -- (10);
		\draw[->,>=latex] (8) -- (9);
		\draw[->,>=latex] (10) -- (11);
		\draw[->,>=latex] (8) -- (11);
		\draw[->,>=latex] (9) -- (11);
		\draw[->,>=latex] (4) -- (11);
		\draw[->,>=latex] (5) -- (11);
		\draw[->,>=latex] (6) -- (8);
		\draw[->,>=latex] (8) to[out=45,in=135] (11);
		\draw (-0.5,1) node {0.4};
		\draw (0.5,3) node {0.3};
		\draw (1.6,3) node {0.3};
		\draw (1,0.3) node {0.3};
		\draw (3,0.3) node {0.1};
		\draw (5,0.3) node {0.1};
		\draw (7,0.3) node {0.3};
		\draw (3,2.3) node {0.2};
		\draw (5.1,2.3) node {0.2};
		\draw (1.5,1) node {0.1};
		\draw (3.5,1) node {0.2};
		\draw (5.5,1) node {0.1};
		\draw (5,4.3) node {0.7};
		\draw (5,5.7) node {2.8};
		\draw (5.2,3) node {0.5};
		\draw (6.5,3) node {0.3};
		\draw (7.2,1.5) node {0.9};
		\draw (8.5,1) node {2.6};
		\draw (3.5,3) node {0.6};
		\end{tikzpicture}
	\end{center}
	\caption{Toy network labeled with link travel times}
	\label{toynetwork}
\end{figure}

Figure \ref{toynetwork} displays a network with one OD pair connected by a set of feasible paths $\mathcal{U}$. We study estimation results for synthetic data of trajectories on this toy network. This data is generated by simulation assuming that the true utility specification is given by $$u_a=\beta_{T}T_a+\beta_{LC}LC_a + \epsilon_a,$$
where $T_a$ is the travel time on arc $a$, and $LC_a$ is a constant equal to 1, with $\beta_T=-2.00$ and $\beta_{LC}=-0.01$. Travel time for each link is given in Figure \ref{toynetwork}. The travelers are also assumed to consider every possible path in $\mathcal{U}$, such that any trajectory may be observed.

We compare the ability of the PL model versus that of the RL model to recover the true parameter values. To do so, we estimate four path-based models based on different choice sets $C \subseteq \mathcal{U}$. Table \ref{tabchoiceset} displays the paths contained in each choice set, noting that choice set $C^4$ contains all 15 paths and is equivalent to $\mathcal{U}$. For each observation, the chosen path is added to the choice set if not already present. The last column displays the utility of each path based on the given specification, obtained by summing the link utilities.

\begin{table}
\begin{center}
\begin{tabular}{clccccc}
Path $\sigma$ &Nodes& $C^1$ & $C^2$& $C^3$ &$v_{\sigma}$\\
\midrule
1&$1-3-4-9-12-13-17$ & y & y & y & $-2.3$\\
2&$1-3-4-7-10-17$ & y & y & y & $-2.6$\\
3&$1-3-4-9-10-17$ & y & y & y & $-2.6$\\
4&$1-3-4-7-10-13-17$ & y & y & y & $-2.7$\\
5&$1-3-4-9-10-13-17$ & y & y & y & $-2.7$\\
\hline
6&$1-3-6-16-17$ & n & y & y & $-3.1$\\
7&$1-3-4-9-12-17$ & n & y & y & $-3.2$\\
8&$1-3-4-7-16-17$ & n & y & y & $-3.2$\\
\hline
9&$1-3-6-16-10-17$ & n & n & y & $-4.0$\\
10&$1-3-4-7-16-10-17$ & n & n & y & $-4.1$\\
11&$1-3-6-16-10-13-17$ & n & n & y & $-4.1$\\
12&$1-3-4-7-16-10-13-17$ & n & n & y & $-4.2$\\
\hline
13&$1-3-4-9-12-15-17$ & n & n & n & $-7.3$\\
14&$1-3-6-16-17$ & n & n & n & $-7.3$\\
15&$1-3-4-7-16-17$ & n & n & n & $-7.4$\\
\end{tabular}
\caption{Paths contained in each restricted choice set}
 \label{tabchoiceset}
\end{center}
\end{table}

Results are shown in Table \ref{example1table}. In the cases where the choice set fails to include several relevant alternatives ($C^1$ and $C^2$), the estimation algorithm for the PL model does not converge, and the parameter values obtained are significantly different from the true ones. The fact that the algorithm does not converge may seem counter-intuitive at first, but it is in fact due to i) the lack of variance in attributes of the paths in these choice sets, ii) the omission in the choice set of paths 6 to 12, which are chosen relatively often in the data, but only added to $C$ when corresponding to the observed path. As a result, when such paths are present in the choice set, the data reports that they are selected $100\%$ of the time, which cannot be reconciled with the explanatory variables present in the utility specification.

The only case where the PL model recovers the true parameter values based on a restricted choice set is with $C^3$, which contains almost the same paths as $\mathcal{U}$ but for three paths. The estimates have slightly lower variance when all alternatives are included with choice set $C^4$, and the RL model obtains equivalent results (the slight difference may be due to different implementations of the optimization algorithm). In accordance with several other studies, we conclude from this experiment that the PL model may not recover the true utility function when the choice set fails to include several relevant alternatives.
	 
 \begin{table}
 	\begin{center}
 		\begin{tabular}{lrr}
 			Model & $\beta_{T}$ & $\beta_{LC}$\\
 			\midrule
 			$PL(C^1)$ & 3.35 (0.59) & $-0.64$ (0.32)\\
 			$PL(C^2)$ & 2.00 (0.37) & 0.64 (0.10)\\
 			$PL(C^3)$ & $-2.06$ (0.17)& $-0.14$ (0.07)\\
            $PL(C^4)$ & $-2.15$ (0.16)& $-0.15$ (0.07)\\
            \hline
            $RL$ & $-2.15$ (0.15) & $-0.14$ (0.07)\\
 		\end{tabular}
 		\caption{Estimation results of different models on synthetic data generated under the assumption that the true choice set is $\mathcal{U}$}
 		\label{example1table}
 	\end{center}
 \end{table}
 
 Certain studies argue that the assumption that users consider any path in $\mathcal{U}$ is behaviorally unrealistic, and inquire what would happen if the data reflects instead the possibility that users do restrict their consideration set. In order to shed light on this question, we study a second sample of synthetic data, where observed trajectories include only paths 1 to 12, generated under the assumption that paths 13 to 15 are not considered by travelers due to their highly negative utility.
 
 In Table \ref{example2table}, we report the estimation results of the same models on this new data. It shows that although the RL model considers more paths than the true choice set, it still recovers the true parameter values while the models based on choice sets $C^1$ and $C^2$ do not. This second experiment suggests that restricting the choice set without evidence regarding what alternatives are truly considered is potentially harmful, while considering a larger set including ``irrelevant'' alternatives does not interfere with estimation results in this case.
     
\begin{table}
 	\begin{center}
 		\begin{tabular}{lrr}
 			Model & $\beta_{T}$ & $\beta_{LC}$\\
 			\midrule
 			$PL(C^1)$ & 3.00 (0.47) & $-0.59$ (0.26)\\
 			$PL(C^2)$ & 3.00 (0.49) & 0.93 (0.13)\\
 			$PL(C^3)$ & $-1.94$ (0.17)& $-0.07$ (0.07)\\
            $PL(C^4)$ & $-2.04$ (0.16)& $-0.07$ (0.07)\\
            \hline
            $RL$ & $-2.04$ (0.15) & $-0.07$ (0.07)\\
 		\end{tabular}
 		\caption{Estimation results of different models on synthetic data generated under the assumption that the true choice set is $C^3$}
 		\label{example2table}
 	\end{center}
 \end{table} 
 
Finally, we note that \cite{frejinger2009} provide a method to correct parameter estimates of path-based models. However, while this leads to consistent estimates, there is no method to consistently predict path choice probabilities according to the estimated model. Indeed, as the next examples highlight, predictions vary significantly depending on the definition of the choice set.
 
\subsection{Examples of prediction}
\label{secpredict}
In general, predicting choices from discrete choice models for a certain demand requires knowing the utility function $v_n$ and choice sets $C_n$ of the decision makers $n$, on which the probability distribution depends. This is in theory more complex when the utility function $v_n$ depends on socio-economic characteristics of individuals $n$. However, in the following, we make the assumption that the utility function is not individual specific and depends only on attributes of network links. 
     
%When the choice set of paths $C_n$ for an individual $n$ traveling between an OD pair is known and can be enumerated, predicting choices from the logit model is straightforward. Indeed the expected flow on a given path $\sigma$ is simply equal to the fraction of the demand choosing $\sigma$ according to $P(\sigma | C_n)$. However since the true choice set $C_n$ is in general unknown, common practice in the literature consists in sampling a set of alternatives for each OD pair including as many relevant routes as possible and to use this choice set for prediction.
     
\subsubsection{Link flows}
\label{secexample2}
Predicting link flows in the network is a typical application of path choice models, of importance in e.g., stochastic user equilibrium models. Link flows represent the amount of individuals (or other unit) on each arc of the network corresponding to loading a certain OD demand.

Two methods exist to predict expected flows with the RL model, none of which require to enumerate choice sets of paths. The first method consists in sampling paths link by link for each individual using link choice probabilities in (\ref{equlinkprob}). The second method allows to compute expected link flows without resorting to simulation. According to the Markov property of the model, \cite{baillon2008markovian}, proved that destination-specific link flows $f^d$ are obtained by solving the linear system
     
     \begin{equation}
     \label{systemlinkflows}
     (I-P^{d^T})^{-1}f^d = g^d,
     \end{equation}
     where $g^d$ is the demand vector from all origins to destination $d$.

In this example, we predict link flows for the network in Figure \ref{toynetwork}, assuming a demand of 100 for the single OD pair and the same utility specification as in section \ref{secexample1}. We compare the link flows predicted by the RL model and the three PL models based on different restricted choice sets $C^1$, $C^2$ and $C^3$. In each case, the expected flow on a given path $\sigma$ is equal to the fraction of the demand choosing $\sigma$ according to $P(\sigma | C_n)$.
     
For the RL model, link flows are obtained by solving (\ref{systemlinkflows}). For the PL models, flows on paths are computed from the path choice probabilities $P(\sigma|C_n)$ for $C_n = C^1, C^2, C^3$. Flows on links are then obtained by summing the flows on all paths traversing each link. Table \ref{tablinkflows} displays the amount of flow on each link according to each model. As expected, we observe that the amount of predicted flow varies greatly between path-based models depending on the chosen choice set. When the choice set size increases, predicted flows tend to be closer to the values forecast by the RL model. A particularity of the RL model is that it predicts non-null flow on every link. However, the amount of flow on links 7, 10 and 16, which belong only to paths with very small choice probabilities, is very close to zero.
     
In reality, it is difficult to judge which model predicts link flows better without being able to compare to observed link counts. However, a crucial remark is that in the absence of any information regarding which paths are truly considered by travelers, the predictions of the PL models are arbitrarily dependent on the choice set. 
%On the other hand, the RL model allows to predict according to the true estimated probability distribution. 
The RL model offers the advantage of computing link flows very efficiently, as only one system of equations must be solved to obtain link flows for all OD pairs with the same destination. In contrast, the PL models require to define a choice set for each OD pair.

An important note is that since the RL model has a non-null choice probability associated to every outgoing link, it predicts some positive cyclic flows in cyclic networks, unlike this example. While this is behaviorally unrealistic, the probabilities associated to paths with loops are in practice extremely small. However, we note that if acyclic link flows are of special interest, by proceeding by simulation it is possible to eliminate paths with cycles when they are sampled from the RL model.
     
\begin{table}
\begin{center}
\begin{tabular}{clrrrr}
Link &Nodes& $PL(C^1)$ & $PL(C^2)$& $PL(C^3)$ & $RL$\\
\midrule
1&o-E & 0.00 & 8.83 & 13.88 & 12.99\\
2&o-A & 100.00 & 91.17 & 86.12 & 87.01\\
3&A-F & 36.92 & 35.75 & 37.10 & 37.39\\
4&A-B & 63.08 & 55.42 & 49.02 & 46.63\\
5&B-C & 26.16 & 27.67 & 24.47 & 25.10\\
6&B-H & 36.92 & 27.75 & 24.55 & 24.53\\
7&C-D & 0.00 & 0.00 & 0.00 & 0.12\\
8&C-d & 0.00 & 8.00 & 7.07 & 6.77\\
9&C-I & 26.16 & 19.67 & 17.40 & 18.21\\
10&D-d & 0.00 & 0.00 & 0.00 & 0.12\\
11&E-G & 0.00 & 8.83 & 13.88 & 12.99\\
12&F-G & 0.00 & 8.00 & 12.55 & 12.86\\
13&F-H & 36.92 & 27.75 & 24.55 & 24.53\\
14&G-H & 0.00 & 0.00 & 11.54 & 12.04\\
15&G-d1 & 0.00 & 16.83 & 14.89 & 13.60\\
16&G-d2 & 0.00 & 0.00 & 0.00 & 0.20 \\
17&H-I &35.08 & 26.36 & 28.80 & 30.40\\
18&H-d &38.76 & 29.14 & 31.84 & 30.70\\
19&I-d &61.24 & 46.03 & 46.20 & 48.60
\end{tabular}
\caption{Link flows according to each model}
 \label{tablinkflows}
\end{center}
\end{table}

\subsubsection{Accessibility measures}
Accessibility measures are another example of information which can be computed from path choice models. The accessibility is a measure of the overall satisfaction of an individual for the available alternatives, i.e., the existing paths in a network for a given OD pair, and is formally defined as the maximum expected utility of the alternatives. According to the RL model, the accessibility is simply the value function at the origin in (\ref{logsum}). In path-based models there is no notion of value function, and instead the accessibility depends on the generated choice set,
\begin{equation}
    E(\max_{i\in C_n} u_i) = \mu \log \sum_{i\in C_n}e^{\frac{1}{\mu}v_i}.
\end{equation}

In the network of this example, accessibility measures are given in Table \ref{accessibility}. This illustrates that the value of accessibility significantly differs depending on choice set composition, and that as more paths are added to $C_n$ the value predicted by PL models converges to that predicted by the RL model, as asserted by \cite{zimmermann2017bike}. Obtaining a prediction of accessibility which is independent of choice sets is very useful, as it allows to compare changes in accessibility before and after network improvements (e.g. after links are added) without bias. When path-based models are used, reported accessibility measures may be incoherent, e.g., decreasing after network improvements, an issue dubbed the Valencia paradox in \cite{nassir2014choice}.

\begin{table}
\begin{center}
\begin{tabular}{rrrr}
$PL(C^1)$ & $PL(C^2)$& $PL(C^3)$ & $RL$\\
\midrule
-0.9592 & -0.6738 & -0.5512 & -0.5478\\
\end{tabular}
\caption{Accessibility measures according to each model}
\label{accessibility}
\end{center}
\end{table}

\section{Conclusion}
\label{secconclu}
This paper presented a tutorial on recursive discrete choice models for the problem of analyzing and predicting path choices in a network. The goal of path choice models is to identify the cost function representing users' behavior, assuming that individuals act rationally by maximizing some kind of objective function when choosing a path in a network. Such models are useful to provide insights into the motivations and preferences of network users and to make aggregate predictions, for instance in the context of traffic equilibrium models.

%In this tutorial, we presented the state of the art methodology for this problem, namely recursive discrete choice models. This methodology is superior in many respects to the discrete choice models based on paths extensively used in the transportation demand modeling literature. This tutorial achieved two main contributions, which we describe below.

This tutorial achieved three purposes, which we describe below. First, we provided a fresh and broader research context for this problem, which has traditionally been addressed mostly from the angle of econometrics in transportation. Namely, this tutorial draws links between discrete choice modeling and related work in inverse optimization and inverse reinforcement learning, which facilitates a greater conceptual understanding of the recursive models presented in this work. %In particular, we contextualized discrete choice as a method for inverse optimization with noisy data. %and showed that viewing the inner problem as a Markov decision process naturally yields the recursive formulation.
In this context, path choice modeling can be viewed as an inverse shortest path problem with noisy data, which can be solved with path-based discrete choice models, or, when modeling the inner shortest path problem as a Markov decision process, with recursive discrete choice models. 

Second, this tutorial presented an overview of the literature on recursive path choice modeling by summarizing the existing recursive model types and describing their properties, estimation algorithms, and methods to compute the value function. We also reviewed and described applications of the methodology to real networks to provide insights regarding computing times and guidance on the choice of model and algorithm.

Third, we discussed and illustrated the potential and advantages of recursive models through a comparison with the most widely used method in the literature, i.e., path-based discrete choice models. While we do not aim at discussing the validity of the behavioral assumptions between both models, we illustrated through examples that recursive models display mathematical convenience, by yielding consistent parameter estimates and predicting path choices without choice set generation. 
\section*{Acknowledgements}
We are very grateful to Tien Mai for his work developing the code which was used for the examples in this tutorial. This work was partly funded by the National Sciences and Engineering Research Council of Canada, discovery grant 435678-2013.

\bibliographystyle{plainnat_custom}	
\bibliography{Predoc}

\begin{thebibliography}{59}
\providecommand{\natexlab}[1]{#1}
\providecommand{\url}[1]{\texttt{#1}}
\expandafter\ifx\csname urlstyle\endcsname\relax
  \providecommand{\doi}[1]{doi: #1}\else
  \providecommand{\doi}{doi: \begingroup \urlstyle{rm}\Url}\fi

\bibitem[Abbeel and Ng(2004)]{abbeel2004apprenticeship}
Abbeel, P. and Ng, A.~Y.
\newblock Apprenticeship learning via inverse reinforcement learning.
\newblock In \emph{Proceedings of the twenty-first international conference on
  Machine learning}, page~1. ACM, 2004.

\bibitem[Aguirregabiria and Mira(2010)]{aguirregabiria2010dynamic}
Aguirregabiria, V. and Mira, P.
\newblock Dynamic discrete choice structural models: A survey.
\newblock \emph{Journal of Econometrics}, 156\penalty0 (1):\penalty0 38--67,
  2010.

\bibitem[Ahuja and Orlin(2001)]{ahuja2001inverse}
Ahuja, R.~K. and Orlin, J.~B.
\newblock Inverse optimization.
\newblock \emph{Operations Research}, 49\penalty0 (5):\penalty0 771--783, 2001.

\bibitem[Akamatsu(1996)]{akamatsu1996cyclic}
Akamatsu, T.
\newblock Cyclic flows, markov process and stochastic traffic assignment.
\newblock \emph{Transportation Research Part B: Methodological}, 30\penalty0
  (5):\penalty0 369--386, 1996.

\bibitem[Aswani et~al.(2018)Aswani, Shen, and Siddiq]{aswani2018inverse}
Aswani, A., Shen, Z.-J., and Siddiq, A.
\newblock Inverse optimization with noisy data.
\newblock \emph{Operations Research}, 2018.

\bibitem[Baillon and Cominetti(2008)]{baillon2008markovian}
Baillon, J.-B. and Cominetti, R.
\newblock Markovian traffic equilibrium.
\newblock \emph{Mathematical Programming}, 111\penalty0 (1-2):\penalty0 33--56,
  2008.

\bibitem[B{\"a}rmann et~al.(2017)B{\"a}rmann, Pokutta, and
  Schneider]{barmann2017emulating}
B{\"a}rmann, A., Pokutta, S., and Schneider, O.
\newblock Emulating the expert: Inverse optimization through online learning.
\newblock In \emph{International Conference on Machine Learning}, pages
  400--410, 2017.

\bibitem[Bekhor and Toledo(2005)]{bekhor2005investigating}
Bekhor, S. and Toledo, T.
\newblock Investigating path-based solution algorithms to the stochastic user
  equilibrium problem.
\newblock \emph{Transportation Research Part B: Methodological}, 39\penalty0
  (3):\penalty0 279--295, 2005.

\bibitem[Bekhor et~al.(2006)Bekhor, Ben-Akiva, and
  Ramming]{bekhor2006evaluation}
Bekhor, S., Ben-Akiva, M.~E., and Ramming, M.~S.
\newblock Evaluation of choice set generation algorithms for route choice
  models.
\newblock \emph{Annals of Operations Research}, 144\penalty0 (1):\penalty0
  235--247, 2006.

\bibitem[Bell(1995)]{BELL1995287}
Bell, M.~G.
\newblock Alternatives to {D}ial's logit assignment algorithm.
\newblock \emph{Transportation Research Part B: Methodological}, 29\penalty0
  (4):\penalty0 287 -- 295, 1995.

\bibitem[Bellman(1958)]{bellman1958routing}
Bellman, R.
\newblock On a routing problem.
\newblock \emph{Quarterly of applied mathematics}, 16\penalty0 (1):\penalty0
  87--90, 1958.

\bibitem[Bertsekas and Tsitsiklis(1991)]{bertsekas1991analysis}
Bertsekas, D.~P. and Tsitsiklis, J.~N.
\newblock An analysis of stochastic shortest path problems.
\newblock \emph{Mathematics of Operations Research}, 16\penalty0 (3):\penalty0
  580--595, 1991.

\bibitem[Bertsimas et~al.(2015)Bertsimas, Gupta, and
  Paschalidis]{bertsimas2015data}
Bertsimas, D., Gupta, V., and Paschalidis, I.~C.
\newblock Data-driven estimation in equilibrium using inverse optimization.
\newblock \emph{Mathematical Programming}, 153\penalty0 (2):\penalty0 595--633,
  2015.

\bibitem[Bliemer and Bovy(2008)]{bliemer2008impact}
Bliemer, M. and Bovy, P.
\newblock Impact of route choice set on route choice probabilities.
\newblock \emph{Transportation Research Record: Journal of the Transportation
  Research Board}, \penalty0 (2076):\penalty0 10--19, 2008.

\bibitem[Bliemer and Taale(2006)]{bliemer2006route}
Bliemer, M.~C. and Taale, H.
\newblock Route generation and dynamic traffic assignment for large networks.
\newblock In \emph{Proceedings of the First International Symposium on Dynamic
  Traffic Assignment, Leeds, UK}, pages 90--99, 2006.

\bibitem[Bovy(2009)]{bovy2009modelling}
Bovy, P.~H.
\newblock On modelling route choice sets in transportation networks: a
  synthesis.
\newblock \emph{Transport reviews}, 29\penalty0 (1):\penalty0 43--68, 2009.

\bibitem[Brotcorne et~al.(2008)Brotcorne, Labbé, Marcotte, and
  Savard]{BrotEtAl08}
Brotcorne, L., Labbé, M., Marcotte, P., and Savard, G.
\newblock Joint design and pricing on a network.
\newblock \emph{Operations Research}, 56\penalty0 (5):\penalty0 1104--1115,
  2008.

\bibitem[Burton and Toint(1994)]{burton1994use}
Burton, D. and Toint, P.~L.
\newblock On the use of an inverse shortest paths algorithm for recovering
  linearly correlated costs.
\newblock \emph{Mathematical Programming}, 63\penalty0 (1-3):\penalty0 1--22,
  1994.

\bibitem[Burton et~al.(1997)Burton, Pulleyblank, and Toint]{burton1997inverse}
Burton, D., Pulleyblank, W., and Toint, P.~L.
\newblock The inverse shortest paths problem with upper bounds on shortest
  paths costs.
\newblock In \emph{Network Optimization}, pages 156--171. Springer, 1997.

\bibitem[Burton and Toint(1992)]{burton1992instance}
Burton, D. and Toint, P.~L.
\newblock On an instance of the inverse shortest paths problem.
\newblock \emph{Mathematical Programming}, 53\penalty0 (1-3):\penalty0 45--61,
  1992.

\bibitem[Chan et~al.(2018)Chan, Lee, and Terekhov]{chan2018inverse}
Chan, T.~C., Lee, T., and Terekhov, D.
\newblock Inverse optimization: Closed-form solutions, geometry, and goodness
  of fit.
\newblock \emph{Management Science}, 2018.

\bibitem[de~Freitas et~al.(2019)de~Freitas, Becker, Zimmermann, and
  Axhausen]{de2019modelling}
de~Freitas, L.~M., Becker, H., Zimmermann, M., and Axhausen, K.~W.
\newblock Modelling intermodal travel in switzerland: A recursive logit
  approach.
\newblock \emph{Transportation Research Part A: Policy and Practice},
  119:\penalty0 200--213, 2019.

\bibitem[Dial(1971)]{Dial71}
Dial, R.~B.
\newblock A probabilistic multipath traffic assignment model which obviates
  path enumeration.
\newblock \emph{Transportation Research}, 5:\penalty0 83 -- 111, 1971.

\bibitem[Dijkstra(1959)]{dijkstra1959note}
Dijkstra, E.~W.
\newblock A note on two problems in connexion with graphs.
\newblock \emph{Numerische mathematik}, 1\penalty0 (1):\penalty0 269--271,
  1959.

\bibitem[Floyd(1962)]{floyd1962algorithm}
Floyd, R.~W.
\newblock Algorithm 97: shortest path.
\newblock \emph{Communications of the ACM}, 5\penalty0 (6):\penalty0 345, 1962.

\bibitem[Fosgerau et~al.(2013)Fosgerau, Frejinger, and
  Karlstr\"om]{fosgerau2013}
Fosgerau, M., Frejinger, E., and Karlstr\"om, A.
\newblock A link based network route choice model with unrestricted choice set.
\newblock \emph{Transportation Research Part B}, 56:\penalty0 70--80, 2013.

\bibitem[Frejinger(2008)]{frejinger2008route}
Frejinger, E.
\newblock \emph{Route choice analysis: data, models, algorithms and
  applications}.
\newblock PhD thesis, Ecole Polytechnique Federale de Lausanne, 2008.

\bibitem[Frejinger and Bierlaire(2007)]{frejinger2007}
Frejinger, E. and Bierlaire, M.
\newblock Capturing correlation with subnetworks in route choice models.
\newblock \emph{Transportation Research Part B}, 41\penalty0 (3):\penalty0
  363--378, 2007.

\bibitem[Frejinger et~al.(2009)Frejinger, Bierlaire, and
  Ben-Akiva]{frejinger2009}
Frejinger, E., Bierlaire, M., and Ben-Akiva, M.
\newblock Sampling of alternatives for route choice modeling.
\newblock \emph{Transportation Research Part B: Methodological}, 43\penalty0
  (10):\penalty0 984--994, 2009.

\bibitem[Gao et~al.(2010)Gao, Frejinger, and Ben-Akiva]{GaoFrejBenA10}
Gao, S., Frejinger, E., and Ben-Akiva, M.
\newblock Adaptive route choices in risky traffic networks: A prospect theory
  approach.
\newblock \emph{Transportation Research Part C: Emerging Technologies},
  18\penalty0 (5):\penalty0 727 -- 740, 2010.

\bibitem[Gilbert et~al.(2015)Gilbert, Marcotte, and Savard]{GilbertEtAl15}
Gilbert, F., Marcotte, P., and Savard, G.
\newblock A numerical study of the logit network pricing problem.
\newblock \emph{Transportation Science}, 49\penalty0 (3):\penalty0 706--719,
  2015.

\bibitem[Horowitz and Louviere(1995)]{horowitz1995role}
Horowitz, J.~L. and Louviere, J.~J.
\newblock What is the role of consideration sets in choice modeling?
\newblock \emph{International Journal of Research in Marketing}, 12\penalty0
  (1):\penalty0 39--54, 1995.

\bibitem[Huang and Bell(1998)]{HuanBell98}
Huang, H.-J. and Bell, M.~G.
\newblock A study on logit assignment which excludes all cyclic flows.
\newblock \emph{Transportation Research Part B: Methodological}, 32\penalty0
  (6):\penalty0 401 -- 412, 1998.

\bibitem[Keshavarz et~al.(2011)Keshavarz, Wang, and
  Boyd]{keshavarz2011imputing}
Keshavarz, A., Wang, Y., and Boyd, S.
\newblock Imputing a convex objective function.
\newblock In \emph{Intelligent Control (ISIC), 2011 IEEE International
  Symposium on}, pages 613--619. IEEE, 2011.

\bibitem[Magnanti and Wong(1984)]{MagnWong84}
Magnanti, T.~L. and Wong, R.~T.
\newblock Network design and transportation planning: Models and algorithms.
\newblock \emph{Transportation Science}, 18\penalty0 (1):\penalty0 1--55, 1984.

\bibitem[Mai(2016)]{mai2016method}
Mai, T.
\newblock A method of integrating correlation structures for a generalized
  recursive route choice model.
\newblock \emph{Transportation Research Part B: Methodological}, 93:\penalty0
  146--161, 2016.

\bibitem[Mai et~al.(2015)Mai, Fosgerau, and Frejinger]{mai2015nested}
Mai, T., Fosgerau, M., and Frejinger, E.
\newblock A nested recursive logit model for route choice analysis.
\newblock \emph{Transportation Research Part B}, 75:\penalty0 100--112, 2015.

\bibitem[Mai et~al.(2016)Mai, Bastin, and Frejinger]{mai2015decomposition}
Mai, T., Bastin, F., and Frejinger, E.
\newblock A decomposition method for estimating recursive logit based route
  choice models.
\newblock \emph{EURO Journal on Transportation and Logistics}, 7\penalty0
  (3):\penalty0 1--23, 2016.

\bibitem[Mai et~al.(2017)Mai, Bastin, and Frejinger]{mai2017similarities}
Mai, T., Bastin, F., and Frejinger, E.
\newblock On the similarities between random regret minimization and mother
  logit: The case of recursive route choice models.
\newblock \emph{Journal of choice modelling}, 23:\penalty0 21--33, 2017.

\bibitem[Manski(1977)]{Manski1977}
Manski, C.~F.
\newblock The structure of random utility models.
\newblock \emph{Theory and Decision}, 8\penalty0 (3):\penalty0 229--254, 1977.

\bibitem[McFadden(1978)]{mcfadden1978}
McFadden, D.
\newblock Modelling the choice of residential location.
\newblock In \emph{Spatial Interaction Theory and Planning Models}, pages
  75--96. A. Karqvist (Ed.), North-Holland, Amsterdam, 1978.

\bibitem[Nassir et~al.(2014)Nassir, Ziebarth, Sall, and Zorn]{nassir2014choice}
Nassir, N., Ziebarth, J., Sall, E., and Zorn, L.
\newblock Choice set generation algorithm suitable for measuring route choice
  accessibility.
\newblock \emph{Transportation Research Record}, \penalty0 (2430):\penalty0
  170--181, 2014.

\bibitem[Nassir et~al.(2018)Nassir, Hickman, and Ma]{nassir2018strategy}
Nassir, N., Hickman, M., and Ma, Z.-L.
\newblock A strategy-based recursive path choice model for public transit smart
  card data.
\newblock \emph{Transportation Research Part B: Methodological}, 126:\penalty0
  528--548, 2018.

\bibitem[Ng et~al.(2000)Ng, Russell, et~al.]{ng2000algorithms}
Ng, A.~Y., Russell, S.~J., et~al.
\newblock Algorithms for inverse reinforcement learning.
\newblock In \emph{Icml}, volume~1, page~2, 2000.

\bibitem[Nielsen and Jensen(2004)]{nielsen2004learning}
Nielsen, T.~D. and Jensen, F.~V.
\newblock Learning a decision maker's utility function from (possibly)
  inconsistent behavior.
\newblock \emph{Artificial Intelligence}, 160\penalty0 (1-2):\penalty0 53--78,
  2004.

\bibitem[Oyama and Hato(2017)]{oyama2017discounted}
Oyama, Y. and Hato, E.
\newblock A discounted recursive logit model for dynamic gridlock network
  analysis.
\newblock \emph{Transportation Research Part C: Emerging Technologies},
  85:\penalty0 509--527, 2017.

\bibitem[Polychronopoulos and Tsitsiklis(1996)]{polychronopoulos1996stochastic}
Polychronopoulos, G.~H. and Tsitsiklis, J.~N.
\newblock Stochastic shortest path problems with recourse.
\newblock \emph{Networks: An International Journal}, 27\penalty0 (2):\penalty0
  133--143, 1996.

\bibitem[Prato and Bekhor(2007)]{prato2007modeling}
Prato, C. and Bekhor, S.
\newblock Modeling route choice behavior: how relevant is the composition of
  choice set?
\newblock \emph{Transportation Research Record: Journal of the Transportation
  Research Board}, \penalty0 (2003):\penalty0 64--73, 2007.

\bibitem[Prato(2009)]{prato2009route}
Prato, C.~G.
\newblock Route choice modeling: past, present and future research directions.
\newblock \emph{Journal of Choice Modelling}, 2\penalty0 (1):\penalty0 65--100,
  2009.

\bibitem[Ratliff et~al.(2006)Ratliff, Bagnell, and
  Zinkevich]{ratliff2006maximum}
Ratliff, N.~D., Bagnell, J.~A., and Zinkevich, M.~A.
\newblock Maximum margin planning.
\newblock In \emph{Proceedings of the 23rd international conference on Machine
  learning}, pages 729--736. ACM, 2006.

\bibitem[Rust(1994)]{rust1994structural}
Rust, J.
\newblock Structural estimation of markov decision processes.
\newblock \emph{Handbook of econometrics}, 4:\penalty0 3081--3143, 1994.

\bibitem[Su and Judd(2012)]{SuJudd12}
Su, C.-L. and Judd, K.~L.
\newblock Constrained optimization approaches to estimation of structural
  models.
\newblock \emph{Econometrica}, 80\penalty0 (5):\penalty0 2213--2230, 2012.

\bibitem[Vovsha and Bekhor(1998)]{vovsha1998link}
Vovsha, P. and Bekhor, S.
\newblock Link-nested logit model of route choice: overcoming route overlapping
  problem.
\newblock \emph{Transportation Research Record: Journal of the Transportation
  Research Board}, \penalty0 (1645):\penalty0 133--142, 1998.

\bibitem[Yai et~al.(1997)Yai, Iwakura, and Morichi]{yai1997multinomial}
Yai, T., Iwakura, S., and Morichi, S.
\newblock Multinomial probit with structured covariance for route choice
  behavior.
\newblock \emph{Transportation Research Part B: Methodological}, 31\penalty0
  (3):\penalty0 195--207, 1997.

\bibitem[Zhang et~al.(1995)Zhang, Ma, and Yang]{zhang1995column}
Zhang, J., Ma, Z., and Yang, C.
\newblock A column generation method for inverse shortest path problems.
\newblock \emph{Zeitschrift f{\"u}r Operations Research}, 41\penalty0
  (3):\penalty0 347--358, 1995.

\bibitem[Ziebart et~al.(2008)Ziebart, Maas, Bagnell, and
  Dey]{ziebart2008maximum}
Ziebart, B.~D., Maas, A.~L., Bagnell, J.~A., and Dey, A.~K.
\newblock Maximum entropy inverse reinforcement learning.
\newblock In \emph{AAAI}, volume~8, pages 1433--1438. Chicago, IL, USA, 2008.

\bibitem[Zimmermann et~al.(2017)Zimmermann, Mai, and
  Frejinger]{zimmermann2017bike}
Zimmermann, M., Mai, T., and Frejinger, E.
\newblock Bike route choice modeling using {GPS} data without choice sets of
  paths.
\newblock \emph{Transportation Research Part C: Emerging Technologies},
  75:\penalty0 183--196, 2017.

\bibitem[Zimmermann et~al.(2018{\natexlab{a}})Zimmermann, Frejinger, and
  Axhausen]{zimmermann2018transit}
Zimmermann, M., Frejinger, E., and Axhausen, K.
\newblock Multi-modal route choice modeling in a dynamic schedule-based transit
  network.
\newblock \emph{15th International Conference on Travel Behavior Research,
  Santa Barbara, California}, 2018{\natexlab{a}}.

\bibitem[Zimmermann et~al.(2018{\natexlab{b}})Zimmermann, V{\"a}stberg,
  Frejinger, and Karlstr{\"o}m]{zimmermann2018capturing}
Zimmermann, M., V{\"a}stberg, O.~B., Frejinger, E., and Karlstr{\"o}m, A.
\newblock Capturing correlation with a mixed recursive logit model for
  activity-travel scheduling.
\newblock \emph{Transportation Research Part C: Emerging Technologies},
  93:\penalty0 273--291, 2018{\natexlab{b}}.

\end{thebibliography}
\end{document}